\documentclass[a4paper, 11pt, oneside]{article}

\usepackage[a4paper, top=3cm, bottom=3cm, left=3cm, right=3cm]{geometry}
\usepackage{helvet}
\usepackage{parskip}
\usepackage[pdftex]{graphicx}
\usepackage{subcaption}
\usepackage[pdftex]{hyperref}
\usepackage{breakcites}
\usepackage{amsmath}
\usepackage{amssymb}
\usepackage{amsfonts}
\usepackage{listings}
\usepackage{animate}

\usepackage{natbib}
\setcitestyle{authoryear,open={[},close={]}}

\usepackage{dirtytalk}

\usepackage[colorinlistoftodos]{todonotes}

\newcommand{\myname}{Steven Abreu}
\newcommand{\mytitle}{Automated Architecture Design for\\Deep Neural Networks}
\newcommand{\mysupervisor}{Prof. Herbert Jaeger}

\hypersetup{
  pdfauthor = {Steven Abreu},
  pdftitle = {Automated Architecture Design for Deep Neural Networks},
  pdfkeywords = {bachelor thesis, machine learning, deep learning, neural networks, deep neural networks, automated architecture design, forward thinking, cascor, neural architecture search, dynamic learning, evolutionary search, neuroevolution, evolving artificial neural networks, artificial neural networks, supervised machine learning},
  colorlinks = {false},
  linkcolor = {black}
}

\begin{document}
  
  \pagenumbering{roman}
  
  \thispagestyle{empty}

  \vspace{30mm}
  \begin{center}
    \rule[0.5ex]{\linewidth}{2pt}\vspace*{-\baselineskip}\vspace*{3.2pt}
    \rule[0.5ex]{\linewidth}{1pt}\\[\baselineskip]
    \huge
    \textbf{\mytitle}
    \rule[0.5ex]{\linewidth}{1pt}\vspace*{-\baselineskip}\vspace{3.2pt}
    \rule[0.5ex]{\linewidth}{2pt}\\
  \end{center}
  \vspace*{4mm}
  \begin{center}
   \Large by
  \end{center}
  \vspace*{4mm}
  \begin{center}
    \Large
    \textsc{\myname}
  \end{center}
  \vspace*{10mm}
  \begin{center}
    \includegraphics[scale=0.6]{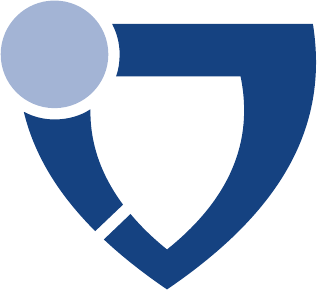}\\
    \vspace*{6mm}
    \large 
    \textsc{Jacobs University Bremen}\\
    \vspace*{10mm}
    \large
    Bachelor Thesis in Computer Science\\
  \end{center}
  \vfill
  \begin{flushright}
    \large
    \begin{tabular}{l}
      \mysupervisor \\[5pt]
      \hline \\[-8pt]
      Bachelor Thesis Supervisor \\
      \\
    \end{tabular}
  \end{flushright}
  \vspace*{8mm}
  \begin{flushleft}
    \large
    Date of Submission: May 17th, 2019 \\
    \rule{\textwidth}{1pt}
  \end{flushleft}
  \begin{center}
    \Large Jacobs University --- Focus Area Mobility
  \end{center}

\newpage
  
  \thispagestyle{empty}

With my signature, I certify that this thesis has been written by me
  using only the indicated resources and materials. Where I have
  presented data and results, the data and results are complete,
  genuine, and have been obtained by me unless otherwise acknowledged;
  where my results derive from computer programs, these computer
  programs have been written by me unless otherwise acknowledged. I
  further confirm that this thesis has not been submitted, either in
  part or as a whole, for any other academic degree at this or another
  institution.

  \vspace{20mm}

  Signature \hfill Place, Date
  
\newpage
  
  \section*{Abstract}
  
%

Machine learning has made tremendous progress in recent years and received large amounts of public attention. Though we are still far from designing a full artificially intelligent agent, machine learning has brought us many applications in which computers solve human learning tasks remarkably well. 
Much of this progress comes from a recent trend within machine learning, called \emph{deep learning}. Deep learning models are responsible for many state-of-the-art applications of machine learning. 

Despite their success, deep learning models are hard to train, very difficult to understand, and often times so complex that training is only possible on very large GPU clusters. Lots of work has been done on enabling neural networks to learn efficiently. However, the design and architecture of such neural networks is often done manually through trial and error and expert knowledge. 
This thesis inspects different approaches, existing and novel, to automate the design of deep feedforward neural networks in an attempt to create less complex models with good performance that take away the burden of deciding on an architecture and make it more efficient to design and train such deep networks.

  \setcounter{tocdepth}{3}
  \tableofcontents
  \clearpage
  
  \pagenumbering{arabic}
  
  \section{Motivation}


\subsection{Relevance of Machine Learning}
\label{ss:rel-ml}

Machine Learning has made tremendous progress in recent years. Although we are not able to replicate human-like intelligence with current state-of-the-art systems, machine learning systems have outperformed humans in some domains. One of the first important milestones has been achieved when DeepBlue defeated the world champion Garry Kasparov in a game of chess in 1997. Machine learning research has been highly active since then and pushed the state-of-the-art in domains like image classification, text classification, localization, question answering, natural language translation and robotics further.


\subsection{Relevance of Deep Learning}
\label{ss:rel-dl}

Many of today's state-of-the-art systems are powered by deep neural networks (see Section \ref{ss:dl}). 
AlphaZero's deep neural network coupled with a reinforcement learning algorithm beat the world champion in Go - a game that was previously believed to be too complex to be played competitively by a machine \citep{alphazero}. 
Deep learning has also been applied to convolutional neural networks - a special kind of neural network architecture that was initially proposed by Yann LeCun \citep{cnn}. 
One of these deep convolutional neural networks, using five layers, has been used to achieve state-of-the-art performance in image classification \citep{imageclass}. 
Overfeat, an eight layer deep convolutional neural network, has been trained on image localization, classification and detection with very competitive results \citep{overfeat}.
Another remarkably complex CNN has been trained with 29 convolutional layers to beat the state of the art in several text classification tasks \citep{textclass}.
Even a complex task that requires coordination between vision and control, such as screwing a cap on a bottle, has been solved competitively using such deep architectures. \cite{robotics} used a deep convolutional neural network to represent policies to solve such robotic tasks. 
Recurrent networks are particularly popular in time series domains. Deep recurrent networks have been trained to achieve state-of-the-art performance in generating captions for given images \citep{captions}. 
Google uses a Long Short Term Memory (LSTM) network to achieve state-of-the-art performance in machine translation \citep{transl}.
Other deep network architectures have been proposed and successfully achieved state-of-the-art performance, such as dynamic memory networks for natural language question answering \citep{question}.

\subsubsection{Inefficiencies of Deep Learning}
\label{ss:dl-inefficiencies}

Evidently, deep neural networks are currently powering many, if not most, state-of-the-art machine learning systems. Many of these deep learning systems train model that are richer than needed and use elaborate regularization techniques to keep the neural network from overfitting on the training data.

Many modern deep learning systems achieve state-of-the-art performance using highly complex models by investing large amounts of GPU power and time as well as feeding the system very large amounts of data. This has been made possible through the recent explosion of computational power as well as through the availability of large amounts of data to train these systems. 

It can be argued that deep learning is inefficient because it trains bigger networks than needed for the function that one desires to learn. This comes at a high expense in the form of computing power, time and the need for larger training datasets.

\subsection{Neural Network Design}
\label{ss:nn-design}



The goal of designing a neural network is manifold. The primary goal is to minimize the neural network's expected loss for the learning task. Because the expected loss cannot always be computed in practice, this goal is often re-defined to minimizing the loss on a set of unseen test data.

Aside from maximizing performance, it is also desirable to minimize the resources needed to train this network. I differentiate between \emph{computational resources} (such as computing power, time and space) and \emph{human resources} (such as time and effort). \\
In my opinion, the goal of minimizing human resources is often overlooked. Many models, especially in deep learning, are designed through trial, error and expert knowledge. This manual design process is rarely interpretable or reproducible and as such, little formal knowledge is gained about the working of neural networks - aside from having a neural network design that may work well for a specific learning task.

In order to avoid the difficulties of defining and assessing the amount of human resources needed for the neural network design process, I am introducing a new goal for the design of neural networks: \emph{level of automaticity}. The level of automaticity in neural network design is inversely proportional to the number of decision that need to be made by a human in the neural network design process.

When dealing with computational resources for neural networks, one might naturally focus on optimizing the amount of computational resources needed during the training process. However, the amount of resources needed for utilizing the neural network in practice are also very important. A neural network is commonly trained once and then used many times once it is trained. The computational resources needed for the utilization of the trained neural network sums up and should be considered when designing a neural network. A good measure is to reduce the model complexity or network size. This goal reduces the computational resources needed for the neural network in practice while simultaneously acting as a regularizer to incentivize neural networks to be smaller - hence prefering simpler models over more complex ones, as Occam's razor states.

To conclude, the goal of designing a neural network is to \textit{maximize performance} (usually by minimizing a chosen loss function on unseen test data), \textit{minimize computational resources} (during training), \textit{maximize the level of automaticity} (by minimizing the amount of decisions that need to be made by a human in the design process), and to \textit{minimize the model's complexity} (e.g. by minimizing the network's size).

  \section{Introduction}

\subsection{Supervised Machine Learning}

In this paper, I will be focusing on \emph{supervised machine learning}. In supervised machine learning, one tries to estimate a function 
$$f: \mathcal{E}_X \mapsto \mathcal{E}_Y$$
where typically $\mathcal{E}_X \subseteq \mathbb{R}^m$ and $\mathcal{E}_Y \subseteq \mathbb{R}^n$, given training data in the form of $(x_i, y_i)_{i=1,..,N}$, with $y_i \approx f(x_i)$. This training data represents existing input-output pairs of the function that is to be estimated.

A machine learning algorithm takes the training data as input and outputs a function estimate $f_{est}$ with $f_{est} \approx f$. The goal of the supervised machine learning task is to minimize a loss function $L$:
$$L: \mathcal{E}_Y \times \mathcal{E}_Y \mapsto \mathbb{R}^{\geq 0}$$
In order to assess a function estimate's accuracy, it should always be assessed on a set of unseen input-output pairs. This is due to \emph{overfitting}, a common phenomenon in machine learning in which a machine learning model memorizes part of the training data which leads to good performance on the training set and (often) bad generalization to unseen patterns. One of the biggest challenges in machine learning is to generalize well. It is trivial to memorize training data and correctly classifying these memorized samples. The challenge lies in correctly classifying previously unseen samples, based on what was seen in the training dataset.

A supervised machine learning problem is specified by labeled training data $(x_i, y_i)_{i=1,..,N}$ with $x_i \in \mathcal{E}_X$, $y_i \in \mathcal{E}_Y$ and a loss function which is to be minimized. Often times, the loss function is not part of the problem statement and instead needs to be defined as part of solving the problem. 

Given training data and the loss function, one needs to decide on a candidate set $\mathcal{C}$ of functions that will be considered when estimating the function $f$.

The learning algorithm $\mathcal{L}$ is an effective procedure to choose one or more particular functions as an estimate for the given function estimation task, minimizing the loss function in some way:
$$\mathcal{L}(\mathcal{C}, L, (x_i, y_i)_{i=1,..N}) \in \mathcal{C}$$
To summarize, a supervised learning problem is given by a set of labeled data points $(x_i, y_i)_{i=1,..N}$ which one typically calls the training data. The loss function $L$ gives us a measure for how good a prediction is compared to the true target value and it can be included in the problem statement. The supervised learning task is to first decide on a candidate set $C$ of functions that will be considered. Finally, the learning algorithm $\mathcal{L}$ gives an effective procedure to choose one function estimate as the solution to the learning problem.

\subsection{Deep Learning}
\label{ss:dl}

Deep learning is a subfield of machine learning that deals with \emph{deep artificial neural networks}. These \emph{artificial neural networks} (ANNs) can represent arbitrarily complex functions (see section \ref{ss:nn-universal}). 

\subsubsection{Artificial Neural Networks}
\label{ss:anns}

An artificial neural network (ANN) (or simply, neural network) consists of a set $V$ of $v = |V|$ processing units, or neurons. Each neuron performs a transfer function of the form
$$y_i = f_i \left( \sum_{j=1}^n w_{ij}x_j - \theta_i \right)$$
where $y_i$ is the output of the neuron, $f_i$ is the activation function (usually a nonlinear function such as the sigmoid function), $x_j$ is the output of neuron $j$, $w_{ij}$ is the connection weight from node $j$ to node $i$ and $\theta_i$ is the bias (or threshold) of the node. Input units are constant, reflecting the function input values. Output units do not forward their output to any other neurons. Units that are neither input nor output units are called hidden units.

The entire network can be described by a directed graph $G = (V, E)$ where the directed edges $E$ are given through a weight matrix $W \in \mathbb{R}^{v \times v}$. Any non-zero entry in the weight matrix at index $(i,j)$, i.e. $w_{ij} \neq 0$ denotes that there is a connection from neuron $j$ to neuron $i$. 

A neural network is defined by its \emph{architecture}, a term that is used in different ways. In this paper, the architecture of a neural network will always refer to the network's node connectivity pattern and the nodes' activation functions.

ANN's can be segmented into feedforward and recurrent networks based on their network topology. An ANN is feedforward if there exists an ordering of neurons such that every neuron is only connected to a neuron further down the ordering. If such an ordering does not exist, then the network is recurrent. In this thesis, I will only be considering feedforward neural networks.

\subsubsection{Feedforward Neural Networks}
\label{ss:ffnn}

A feedforward network can be visualized as a layered network, with layers $L_0$ through $L_K$. The layer $L_0$ is called the input layer and $L_K$ is called the output layer. Intermediate layers are called hidden layers.

One can think of the layers as subsequent feature extractors: the first hidden layer $L_1$ is a feature extractor on the input unit. The second hidden layer $L_2$ is a feature extractor on the first hidden layer - thus a second order feature extractor on the input. The hidden layers can compute increasingly complex features on the input.

\subsubsection{Neural Networks as Universal Function Approximators}
\label{ss:nn-universal}

A classical universal approximation theorem states that standard feedforward neural networks with only one hidden layer using a squashing activation function (a function $\Psi: \mathbb{R} \mapsto [0,1]$ is a squashing function, according to \cite{mlpuniversalapprox}, if it is non-decreasing, $\Psi_{\lambda \rightarrow \infty} (\lambda) = 1$ and $\Psi_{\lambda \rightarrow - \infty} (\lambda) = 0$) can be used to approximate any continuous function on \emph{compact subsets} of $\mathbb{R}^n$ with any desired non-zero amount of error \citep{mlpuniversalapprox}. The only requirement is that the network must have \emph{sufficiently many} units in its hidden layer. 

\begin{figure}[h]
  \center
  \includegraphics[width=0.8\textwidth]{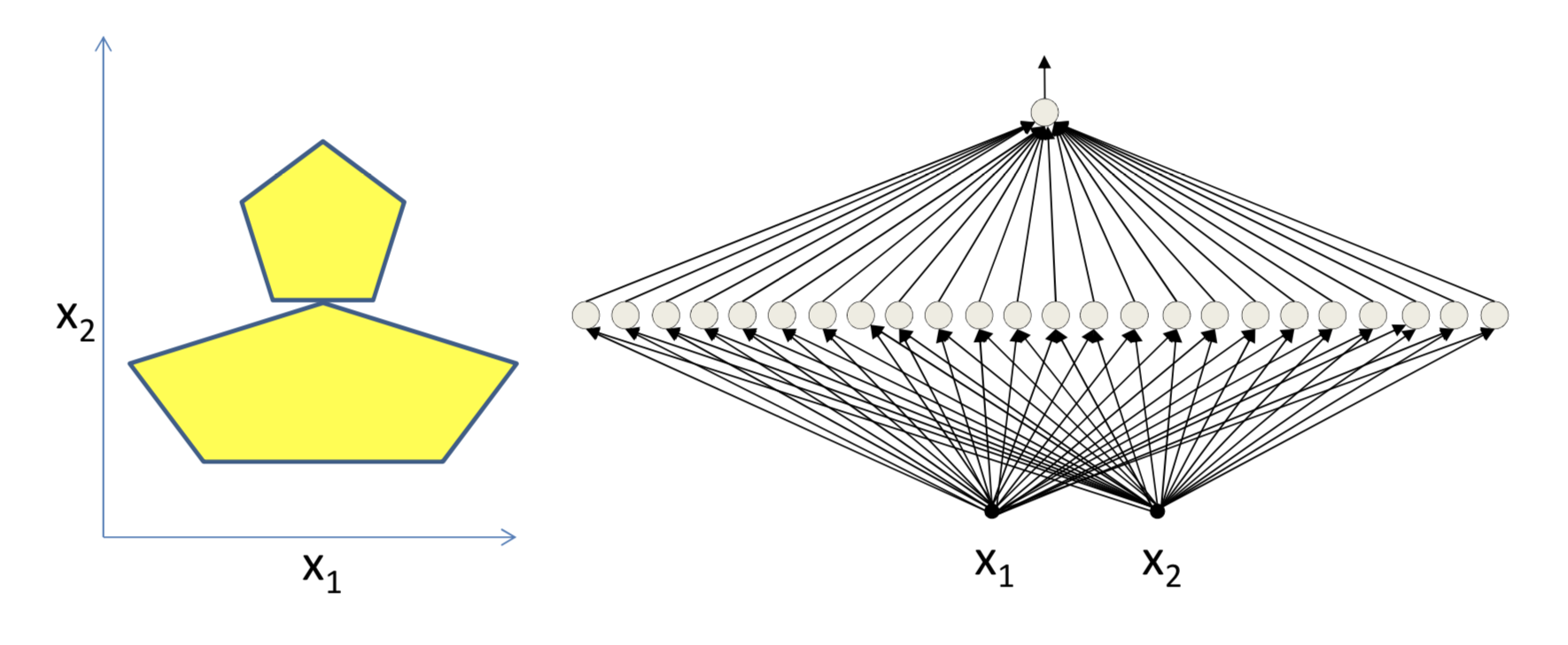}
  \caption{Binary classification problem. Yellow area is one class, everything else is the other class. Right is the shallow neural network that should represent the classification function. Figure taken from Bhiksha Raj's lecture slides in CMU's '11-785 Introduction to Deep Learning'.}
  \label{fig:bin-classification-problem}
\end{figure}

A simple example can demonstrate this universal approximation theorem for neural networks. Consider the binary classification problem in Figure \ref{fig:bin-classification-problem} of the kind $f: [0,1]^2 \rightarrow \{0,1\}$. The function solving this classification problem can be represented using an MLP. As stated by the universal approximation theorem, one can approximate this function to arbitrary precision using an MLP with one hidden layer. 

The difficulty in representing the desired classification function is that the classification is split into two separate, disconnected decision regions. Representing either one of these shapes is trivial. One can add one neuron per side of the polygon which acts as a feature detector to detect the decision boundary represented by this side of the polygon. One can then add a bias into the hidden layer with a value of $b_h = -N$ ($N$ is the number of sides of the polygon), use a relu-activated output unit and one has built a simple neural network which returns $1$ iff all hidden neurons fire, i.e. when the point lies within the boundary of every side of the polygon, i.e. when the point lies within the polygon. 

\begin{figure}[h]
  \center
  \begin{subfigure}[b]{0.2\textwidth}
    \includegraphics[width=\textwidth]{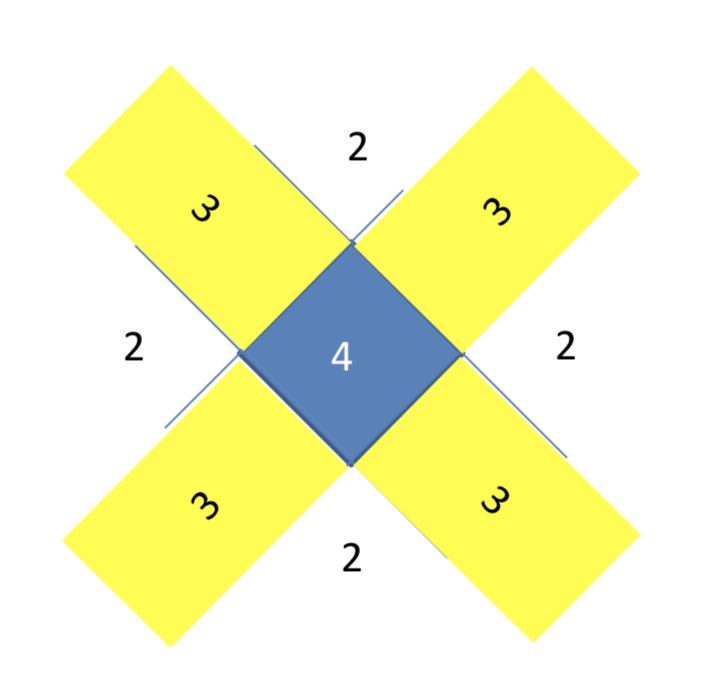}
    \caption{Decision boundary for a square}
    \label{fig:decision-boundary-4}
  \end{subfigure}
  ~~
  \begin{subfigure}[b]{0.2\textwidth}
    \includegraphics[width=\textwidth]{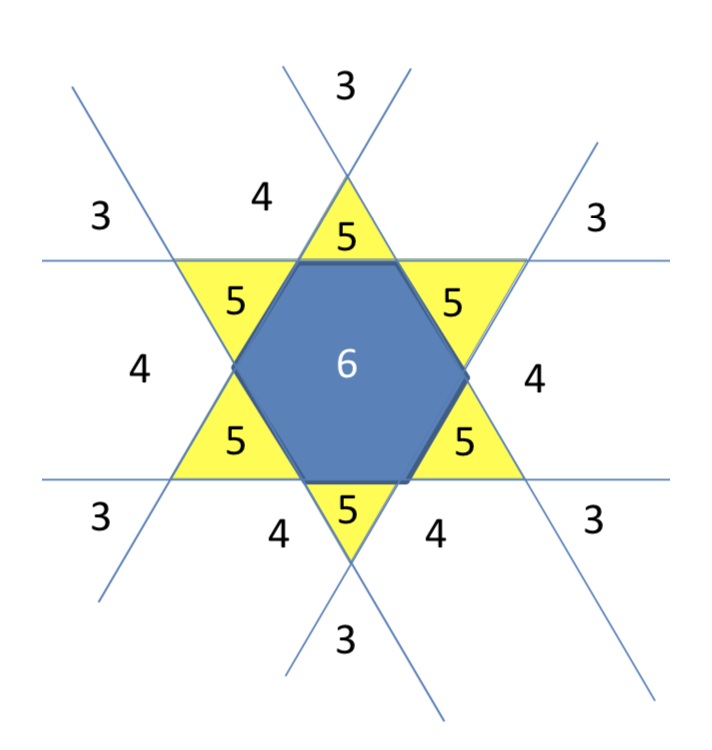}
    \caption{Decision boundary for a hexagon}
    \label{fig:decision-boundary-6}
  \end{subfigure}
  ~~ 
  \begin{subfigure}[b]{0.2\textwidth}
    \includegraphics[width=\textwidth]{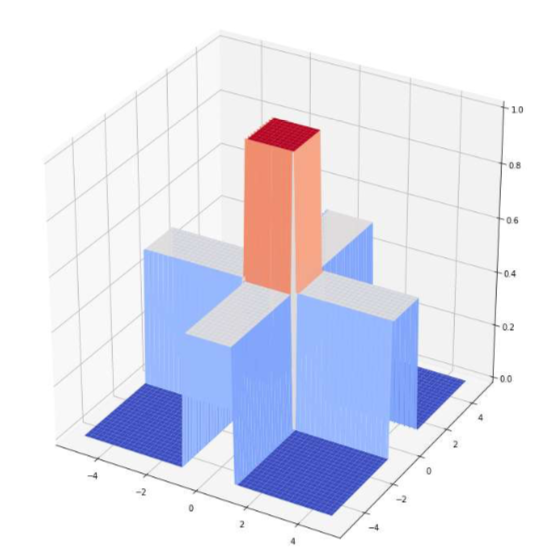}
    \caption{Decision plot for a square}
    \label{fig:decision-plot-4}
  \end{subfigure}
  ~~ 
  \begin{subfigure}[b]{0.2\textwidth}
    \includegraphics[width=\textwidth]{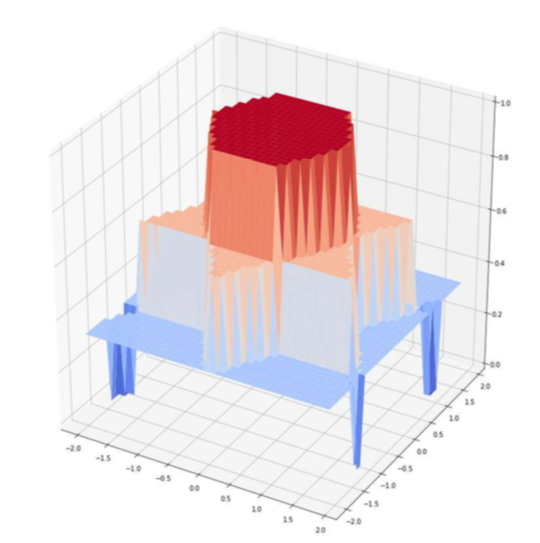}
    \caption{Decision plot for a hexagon}
    \label{fig:decision-plot-6}
  \end{subfigure}
  \caption{Decision plots and boundaries for simple binary classification problems. Figures taken from Bhiksha Raj's lecture slides in CMU's '11-785 Introduction to Deep Learning'.}
\end{figure}

This approach generalizes neither to shapes that are not convex nor to multiple, disconnected shapes. In order to approximate any decision boundary using just one hidden layer, one can use an $n$-sided polygon. Figure \ref{fig:decision-boundary-4} and \ref{fig:decision-boundary-6} show the decision boundaries for a square and a hexagon. A problem arises when the two shapes are close to each other; the areas outside the boundaries add up to values larger or equal to those within the boundaries of each shape. In the plots of Figure \ref{fig:decision-plot-4} and \ref{fig:decision-plot-6}, one can see that the boundaries of the decision regions don't fall off quickly enough and will add up to large values, if there are two or more such shapes in close proximity.

\begin{figure}[h]
  \center 
  \includegraphics[width=0.8\textwidth]{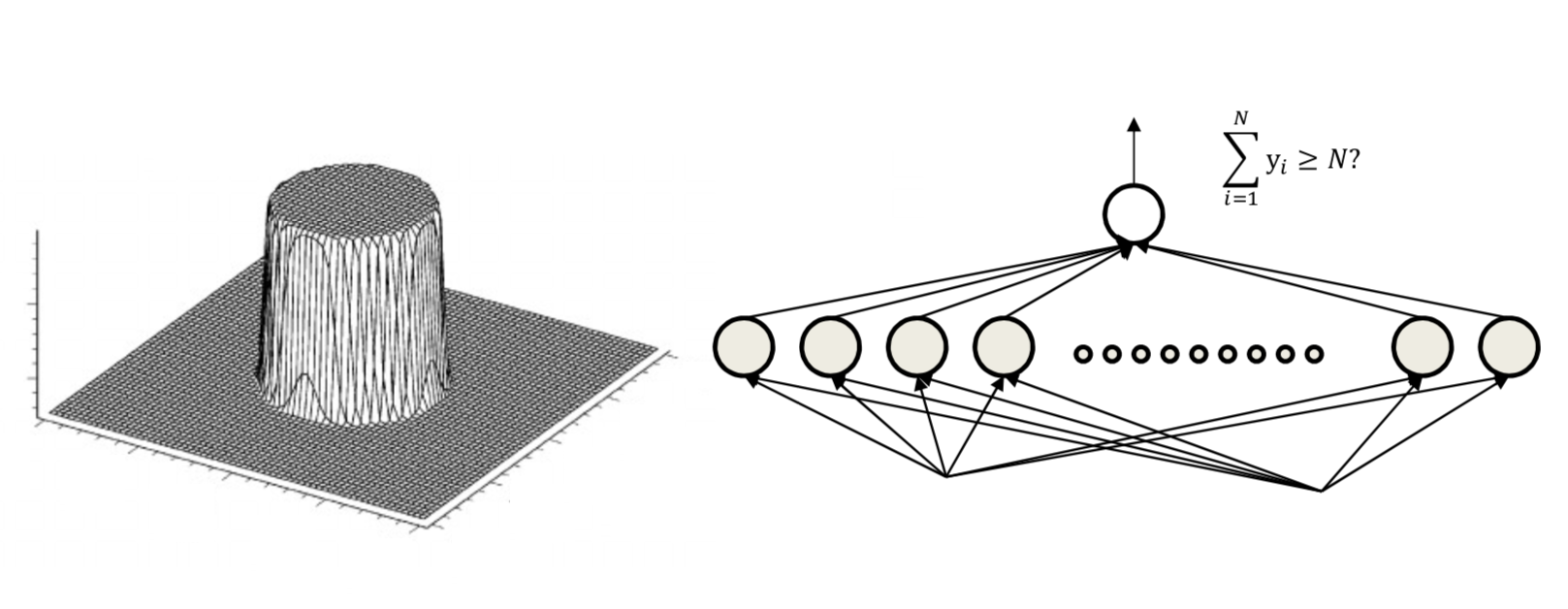}
  \caption{Decision plot and corresponding MLP structure for approximating a circle. Figure taken from Bhiksha Raj's lecture slides in CMU's '11-785 Introduction to Deep Learning'.}
  \label{fig:decision-plot-limit}
\end{figure}

However, as one increases the sides $n$ of the polygon, the boundaries will fall off more quickly. In the limit of $n \rightarrow \infty$, the shape becomes a near perfect cylinder, with value $n$ for the area within the cylinder and $n/2$ outside. Using a bias unit of $b_h=-n/2$, one can turn this into a near-circular shape with value $n/2$ in the shape and value $0$ everywhere else, as shown in Figure \ref{fig:decision-plot-limit}. One can now add multiple near-circles together in the same layer of the neural network. Given this setup, one can now compose an arbitrary figure by fitting it with an arbitrary number of near-circles. The smaller these near-circles, the more accurate this classification problem can be represented by a network. With this setup, it is possible to capture any decision boundary.

This procedure to build a neural network with one hidden layer to build a classifier for arbitrary figures has a problem: the number of hidden units needed to represent this function become arbitrarily high. In this procedure, I have set $n$, the number of hidden units to represent a circle to be very large and I am using many of these circles to represent the entire function. This will result in a very (\textit{very}) large number of units in the hidden layer. 

This is a general phenomenon: even though a network with just one hidden layer can represent any function (with some restrictions, see above) to arbitrary precision, the number of units in this hidden layer often becomes intractably large. Learning algorithms often fail to learn complicated functions correctly without overfitting the training data in such "shallow" networks. 

\subsubsection{Relevance of Depth in Neural Networks}
\label{ss:dl-depth}

\begin{figure}[h]
  \center 
  \includegraphics[width=0.8\textwidth]{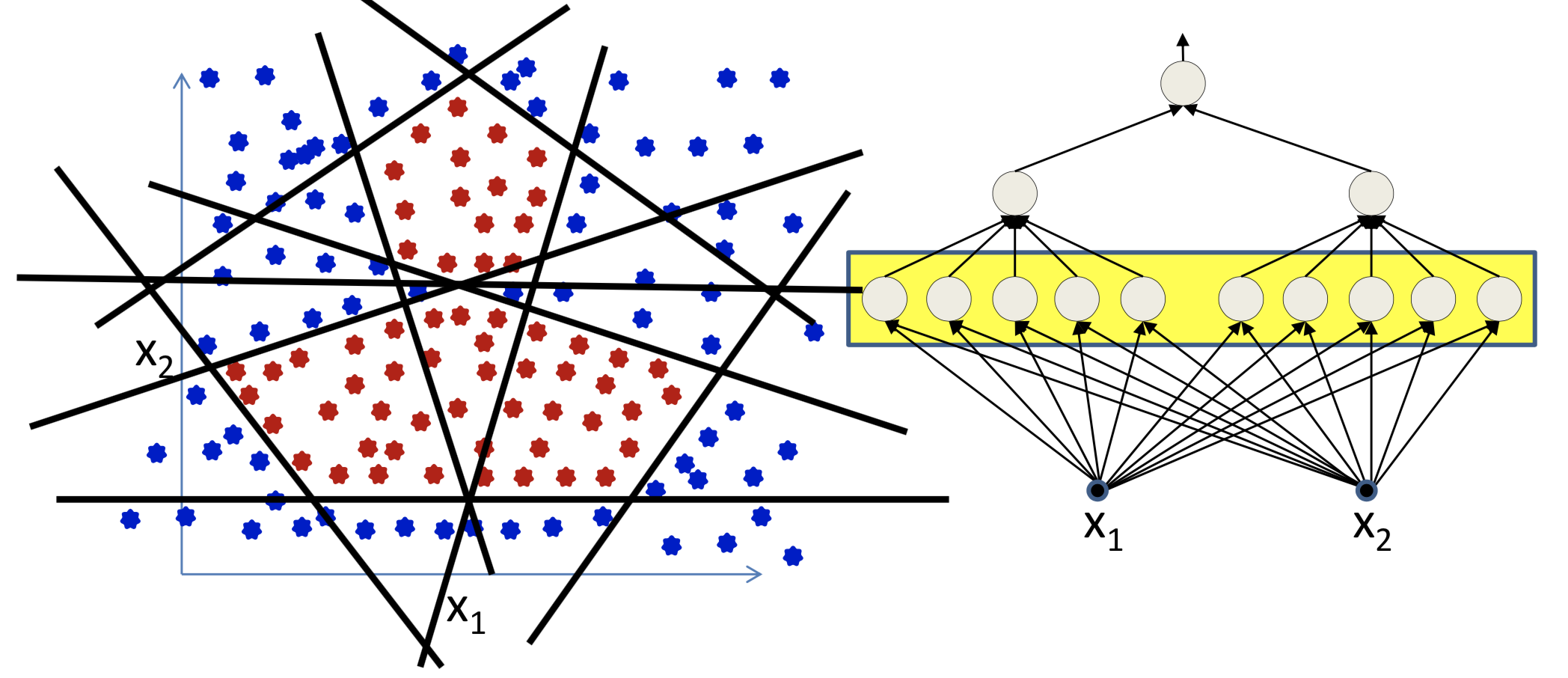}
  \caption{Decision boundary and corresponding two-layer classification network. Figure taken from Bhiksha Raj's lecture slides in CMU's '11-785 Introduction to Deep Learning'.}
  \label{fig:decision-boundary-deep}
\end{figure}

The classification function from Figure \ref{fig:bin-classification-problem} can be built using a smaller network, if one allows for multiple hidden layers. The first layer is a feature detector for every polygon's edge. The second layer will act as an AND gate for every distinct polygon - detecting all those points that lie within all the  polygon's edges. The output layer will then act as an OR gate for all neurons in the second layer, thus detecting all points that lie in \textit{any} of the polygons. With this, one can build a simple network that perfectly represents the desired classification function. The network and decision boundaries are shown in Figure \ref{fig:decision-boundary-deep}.

By adding just one additional layer into the network, the number of hidden neurons has been reduced from $n_{\text{shallow}} \rightarrow \infty$ to $n_{\text{deep}} = 12$. This shows how the depth of a network can increase the resulting model capacity faster than an increase in the number of units in the first hidden layer.

\subsubsection{Advantages of Deeper Neural Networks}
\label{ss:imp-dl}

It is difficult to understand how the depth of an arbitrary neural network influences what kind of functions the network can compute and how well these networks can be trained. Early research has focused on shallow networks and their conclusions cannot be generalized to deeper architectures, such as the universal approximation theorem for networks with one hidden layer \citep{mlpuniversalapprox} or an analysis of a neural network's expressivity based on an analogy to boolean circuits by \cite{maass1994comparison}.

Several measures have been proposed to formalize the notion of model capacity and the complexity of functions which a statistical learning algorithm can represent. One of the most famous such formalization is that of the Vapnik Chervonenkis dimension (VC dimension) \citep{vapnik2015uniform}. 

Recent papers have focused on understanding the benefits of depth in neural networks. The VC dimension as a measure of capacity has been applied to feedforward neural network with piecewise polynomial activation functions, such as relu, to prove that a network's model capacity grows by a factor of $\frac{W}{\log W}$ with depth compared to a similar growth in width \citep{bartlett1999almost}. 

There are examples of functions that a deeper network can express and a more shallow network cannot approximate unless the width is exponential in the dimension of the input (\citep{eldan2016power} and \citep{telgarsky2015representation}). Upper and lower bounds have been established on the network complexity for different numbers of hidden units and activation functions. These show that deep architectures can, with the same number of hidden units, realize maps of higher complexity than shallow architectures \citep{bianchini2014complexity}.

However, the aforementioned papers either do not take into account the depth of modern deep learning models or only present findings for specific choices of weights of a deep neural network.

Using Riemannian geometry and dynamical mean field theory, \cite{poole2016exponential} show that generic deep neural networks can "efficiently compute highly expressive functions in ways that shallow networks cannot" which "quantifies and demonstrates the power of deep neural networks to disentangle curved input manifolds" \citep{poole2016exponential}.

\cite{raghu2017expressive} introduced the notion of a \emph{trajectory}; given two points in the input space $x_0, x_1 \in \mathbb{R}^m$, the trajectory $x(t)$ is a curve parametrized by $t \in [0,1]$ with $x(0)=x_0$ and $x(1)=x_1$. They argue that the trajectory's length serves as a measure of network expressivity. By measuring the trajectory lengths of the input as it is transformed by the neural network, they found that the network's depth increases complexity (given by the trajectory length) of the computed function exponentially, compared to the network's width.

\subsubsection{The Learning Problem in Neural Networks}

A network architecture being able to \textit{approximate} any function does not always mean that a network of that architecture is able to \textit{learn} any function. Whether or not neural network of a fixed architecture can be trained to represent a given function depends on the learning algorithm used.

The learning algorithm needs to find a set of parameters for which the neural network computes the desired function. Given a function, there exists a neural network to represent this function. But even if such an architecture is given, there is no universal algorithm which, given training data, finds the correct set of parameters for this network such that it will also generalize well to unseen data points \citep{deeplearning}.

Finding the optimal neural network architecture for a given learning task is an unsolved problem as well. \cite{zhang2016understanding} argue that most deep learning systems are built on models that are rich enough to memorize the training data.

Hence, in order for a neural network to learn a function from data, it has to learn the network architecture and the parameters of the neural network (connection weights). This is commonly done in sequence but it is also possible to do both simultaneously or iteratively.

  \section{Automated Architecture Design}
\label{ss:aut-arch-des}

Choosing a fitting architecture is a big challenge in deep learning. Choosing an unsuitable architecture can make it impossible to learn the desired function. Choosing an optimal architecture for a learning task is an unsolved problem. Currently, most deep learning systems are designed by experts and the design relies on hyperparameter optimization through a combination of grid search and manual search \citep{bergstra2012random} (see \cite{larochelle2007empirical}, \cite{lecun2012efficient}, and \cite{hinton2012practical}).

This manual design is tedious, computationally expensive, and architecture decisions based on experience and intuition are very difficult to formalize and thus, reuse. Many algorithms have been proposed for the architecture design of neural networks, with varying levels of automaticity. In this thesis, I will be referring to these algorithms as \emph{automated architecture design algorithms}.

Automated architecture algorithms can be broadly segmented into neural network architecture search algorithms (also called neural architecture search, or NAS) and dynamic learning algorithms, both of which are discussed in this section.

\subsection{Neural Architecture Search}

Neural architecture search is a natural choice for the design of neural networks. NAS methods are already outperforming manually designed architectures in image classification and object detection (\citep{Zoph_2018_CVPR} and \citep{real2018regularized}).

\cite{elsken2018neural} propose to categorize NAS algorithms according to three dimensions: search space, search strategy, and performance estimation strategy. The authors describe these as follows.
The search space defines the set of architectures that are considered by the search algorithm. Prior knowledge can be incorporated into the search space, though this may limit the exploration of novel architectures.
The search strategy defines the search algorithm that is used to explore the search space. The search algorithm defines how the exploration-exploitation tradeoff is handled.
The performance estimation strategy defines how the performance of a neural network architecture is assessed. Naively, one may train a neural network architecture but this is object to random fluctuations due to initial random weight initializations, and obviously very computationally expensive.

In this thesis, I will not be considering the search space part of the NAS algorithms. Instead, I will keep the search space constant across all NAS algorithms. I will not go in depth about the performance estimation strategy in the algorithms either, instead using one constant form of constant estimation - training a network architecture once for the same number of epochs (depending on time constraints).

Many search algorithms can be used in NAS algorithms. \cite{elsken2018neural} names random search, Bayesian optimization, evolutionary methods, reinforcement learning, and gradient-based methods. Search algorithms can be divided into adaptive and non-adaptive algorithms, where adaptive search algorithms adapt future searches based on the performance of already tested instances. In this thesis, I will only consider grid search and random search as non-adaptive search algorithms, and evolutionary search as an adaptive search algorithm.

For the following discussion, let $\mathcal{A}$ be the set of all possible neural network architectures and $\mathcal{A}' \subseteq \mathcal{A}$ be the search space defined for the NAS algorithm - a subset of all possible architectures. 

\subsubsection{Non-Adaptive Search - Grid and Random Search}
\label{sss:nonadaptive-search}

The simplest way to automatically design a neural network's architecture may be to simply try different architectures from a defined subset of all possible neural network architectures and choose the one that performs the best. One chooses elements $a_i \in \mathcal{A}'$, tests these individual architectures and chooses the one that performs the best. The performance is usually measured through evaluation on an unseen testing set or through a cross validation procedure - a technique which artificially splits the training data into training and validation data and uses the unseen validation data to evaluate the model's performance.

The two most widely known search algorithms that are frequently used for hyperparameter optimization (which includes architecture search) are \emph{grid search} and \emph{random search}. Naive grid search performs an exhaustive, enumerated search within the chosen subset $\mathcal{A}'$ of possible architectures - where one needs to also specify some kind of step size, a discretization scheme which determines how "fine" the search within the architecture subspace should be. Adaptive grid search algorithms use adaptive grid sizes and are not exhaustive. Random search does not need a discretization scheme, it chooses elements from $\mathcal{A}'$ at random in each iteration. Both grid and random search are non-adaptive algorithms: they do not vary the course of the experiment by considering the performance of already tested instances \citep{bergstra2012random}. \cite{larochelle2007empirical} finds that, in the case of a 32-dimensional search problem of deep belief network optimization, random search was not as good as the sequential combination of manual and grid search from an expert because the efficiency of sequential optimization overcame the inefficiency of the grid search employed at every step \citep{bergstra2012random}. \cite{bergstra2012random} concludes that sequential, adaptive algorithms should be considered in future work and random search should be used as a performance baseline.

\subsubsection{Adaptive Search - Evolutionary Search}

In the past three decades, lots of research has been done on genetic algorithms and artificial neural networks. The two areas of research have also been combined and I shall refer to this combination as evolving artificial neural networks (EANN), based on a literature review by \cite{evolvinganns}. Evolutionary algorithms have been applied to artificial neural networks to evolve connection weights, architectures, learning rules, or any combination of these three. These EANN's can be viewed as an adaptive system that is able to learn from data as well as evolve (adapt) its architecture and learning rules - without human interaction. 

Evolutionary algorithms are population based search algorithms which are derived from the principles of natural evolution. They are very useful in complex domains with many local optima, as is the case in learning the parameters of a neural network \citep{lossmln}. They do not require gradient information which can be a computational advantage as the gradients for neural network weights can be quite expensive to compute, especially so in deep networks and recurrent networks. The simultaneous evolution of connection weights and network architecture can be seen as a fully automated ANN design. The evolution of learning rules can be seen as a way of "learning how to learn". In this paper, I will be focusing on the evolution of neural network architectures, staying independent of the algorithm that is used to optimize connection weights.

The two key issues in the design of an evolutionary algorithm are the representation and the search operators. The architecture of a neural network is defined by its nodes, their connectivity and each node's transfer function. The architecture can be encoded as a string in a multitude of ways, which will not be discussed in detail here. 

A general cycle for the evolution of network architectures has been proposed by \cite{evolvinganns}:

\begin{enumerate}
  \item Decode each individual in the current generation into an architecture.
  \item Train each ANN in the same way, using $n$ distinct random initializations.
  \item Compute the fitness of each architecture according to the averaged training results.
  \item Select parents from the population based on their fitness.
  \item Apply search operators to parents and generate offspring to form the next generation.
\end{enumerate}


It is apparent that the performance of an EANN depends on the encoding scheme of the architecture, the definition of the fitness function, and the search operators applied to the parents to generate offspring. There will be some residual noise in the process due to the stochastic nature of ANN training. Hence, one should view the computed fitness as a heuristic value, an approximation, for the true fitness value of an architecture. The larger the number $n$ of different random initializations that are run for each architecture, the more accurate training results (and thus, the fitness computation) becomes. However, increasing $n$ leads to a large increase in time needed for each iteration of the evolutionary algorithm. 

\subsection{Dynamic Learning}

\begin{figure}[h]
  \center 
  \includegraphics[width=0.6\textwidth]{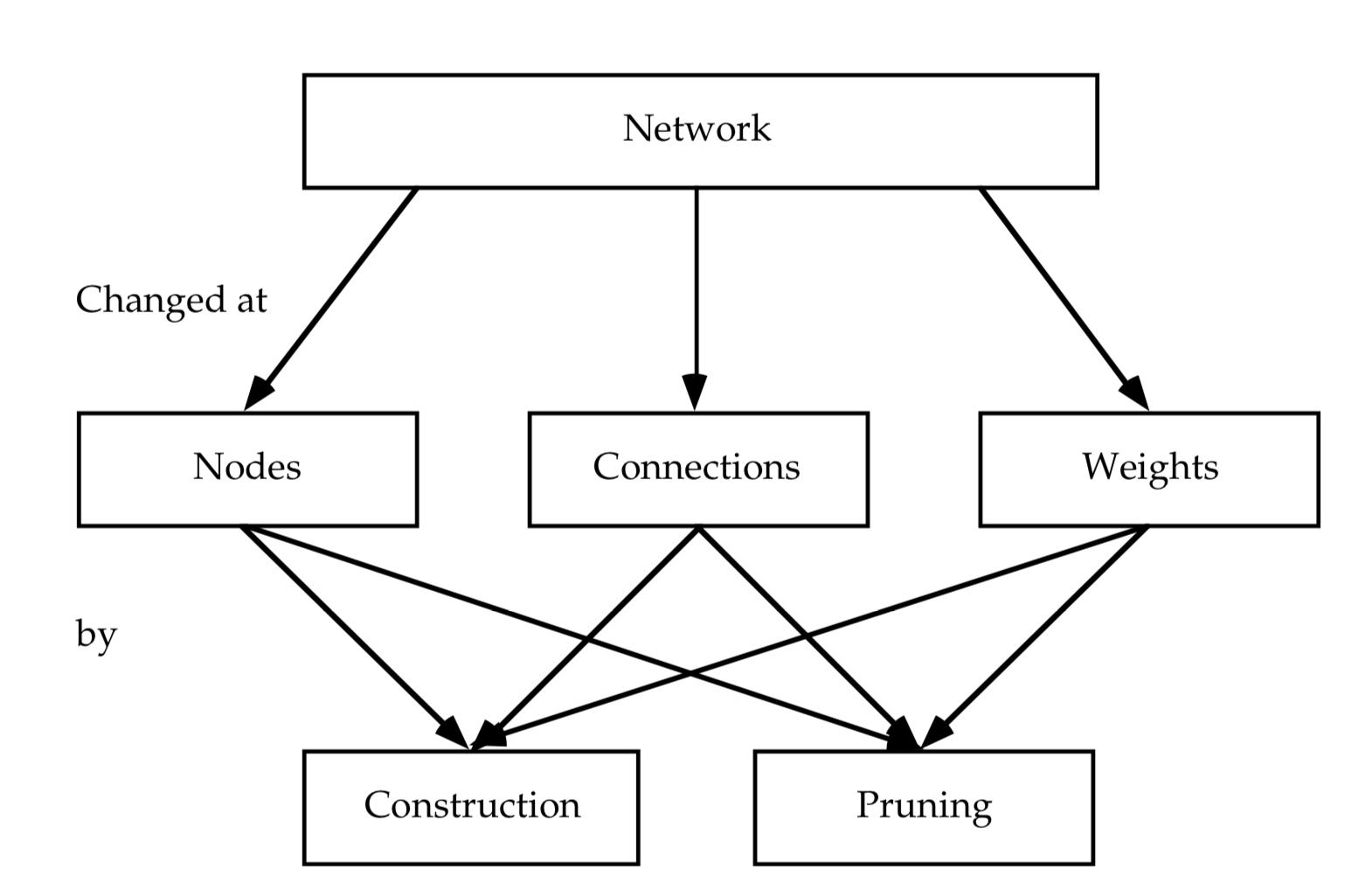}
  \caption{Possible network topology changes, taken from \cite{waugh1994dynamic}}
  \label{fig:net-top-changes}
\end{figure}

Dynamic learning algorithms in neural networks are algorithms that modify a neural network's hyperparameters and topology (here, I focus on the network architecture) dynamically as part of the learning algorithm, during training. These approaches present the opportunity to develop optimal network architectures that generalize well \citep{waugh1994dynamic}. The network architecture can be modified during training by adding complexity to the network or by removing complexity from the network. The former is called a constructive algorithm, the latter a destructive algorithm. Naturally, the two can be combined into an algorithm that can increase and decrease the network's complexity as needed, in so-called combined dynamic learning algorithms. These changes can affect the nodes, connections or weights of the network - a good overview of possible network changes is given by \cite{waugh1994dynamic}, see Figure \ref{fig:net-top-changes}.

\subsubsection{Regularization Methods}

Before moving on to dynamic learning algorithms, it is necessary to clear up the classification of these dynamic learning algorithms and clarify some underlying terminology. The set of destructive dynamic learning algorithms intersects with the set of so-called regularization methods in neural networks. The origin of this confusion is the definition of dynamic learning algorithms. \cite{waugh1994dynamic} defines dynamic learning algorithms to change either the nodes, connections, or \emph{weights} of the neural network. If we continue with this definition, we will include all algorithms that reduce the values of connections weights in the set of destructive dynamic learning, which includes regularization methods.

Regularization methods penalize higher connection weights in the loss function (as a result, connection weights are reduced in value). Regularization is based on Occam's razor which states that the simplest explanation is more likely to be correct than more complex explanations. Regularization penalizes such complex explanations (by reducing the connection weights' values) in order to simplify the resulting model. 

Regularization methods include weight decay, in which a term is added to the loss function which penalizes large weights, and dropout, which is explained in Section \ref{ss:dropout}. For completeness, I will cover these techniques as instances of dynamic learning, however I will not run any experiments on these regularization methods as the goal of this thesis is to inspect methods to automate the \emph{architecture} design, for which the modification of connection weights is not relevant.

\subsubsection{Destructive Dynamic Learning}

In destructive dynamic learning, one starts with a network architecture that is larger than needed and reduces complexity in the network by removing nodes, connections or reducing existing connection weights. 

A key challenge in this destructive approach is the choice of starting network. As opposed to a minimal network - which could simply be a network without any hidden units - it is difficult to define a "maximal" network because there is no upper bound on the network size \citep{waugh1994dynamic}. A simple solution would be to choose a fully connected network with $K$ layers, where $K$ is dependent on the learning task. 

An important downside to the use of destructive algorithms is the computational cost. Starting with a very large network and then cutting it down in size leads to many redundant computations on the large network. 

Most approaches to destructive dynamic learning that modify the nodes and connections (rather than just the connection weights) are concerned with the pruning of hidden nodes. The general approach is to train a network that is larger than needed and prune parts of the network that are not essential. \cite{reed1993pruning} suggests that most pruning algorithms can be divided into two groups; algorithms that estimate the sensitivity of the loss function with respect to the removal of an element and then removes those elements with the smallest effect on the loss function, and those that add terms to the objective function that rewards the network for choosing the most efficient solution - such as weight decay. I shall refer to those two groups of algorithms as sensitivity calculation methods and penalty-term methods, respectively - as proposed by \cite{waugh1994dynamic}.

Other algorithms have been proposed but will not be included in this thesis for brevity reasons (most notably, principal components pruning \citep{levin1994fast} and soft weight-sharing as a more complex Penalty-Term method \citep{nowlan1992simplifying}).

\paragraph{Dropout}\mbox{}\\
\label{ss:dropout}

This section follows \cite{srivastava2014dropout}. Dropout refers to a way of regularizing a neural network by randomly "dropping out" entire nodes with a certain probability $p$ in each layer of the network. At the end of training, each node's outgoing weights are then multiplied with its probability $p$ of being dropped out. As the networks connection weights are multiplied with a certain probability value $p$, where $p \in [0,1]$, one can consider this technique a kind of connection weight pruning and thus, in the following, I will consider dropout to be a destructive algorithm.

Intuitively, dropout drives hidden units in a network to work with different combinations of other hidden units, essentially driving the units to build useful features without relying on other units. Dropout can be interpreted as a stochastic regularization technique that works by introducing noise to its units. 

One can also view this "dropping out" in a different way. If the network has $n$ nodes (excluding output notes), dropout can either include or not include this node. This leads to a total of $2^n$ different network configurations. At each step during training, one of these network configurations is chosen and the weights are optimized using some gradient descent method. The entire training can hence be seen as training not just one network but all possible $2^n$ network architectures. 
In order to get an ideal prediction from a flexible-sized model such as a neural network, one should average over the predictions of all possible settings of the parameters, weighing each setting by its posterior probability given the training data. This procedure quickly becomes intractable. In essence, dropout is a technique that can combine exponentially (exponential in the number of nodes) many different neural networks efficiently.

Due to this model combination, dropout is reported to take 2-3 times longer to train than a standard neural network without dropout. This makes dropout an effective algorithm that deals with a trade-off between overfitting and training time.

To conclude, dropout can be seen as both a regularization technique and a form of model averaging. It works remarkably well in practice. \cite{srivastava2014dropout} report large improvements across all architectures in an extensive empirical study. The overall architecture is not changed, as the pruning happens only in terms of the magnitude of the connection weights.

\paragraph{Penalty-Term Pruning through Weight Decay}\mbox{}\\
\label{ss:weight-decay}

Weight decay is the best-known regularization technique that is frequently used in deep learning applications. It works by penalizing network complexity in the loss function, through some complexity measure that is added into the loss function - such as the number of free parameters or the magnitude of connection weights. \cite{krogh1992simple} show that weight decay can improve generalization of a neural network by suppressing irrelevant components of the weight vector and by suppressing some of the effect of static noise on the targets.

\paragraph{Sensitivity Calculation Pruning}\mbox{}\\

\cite{sietsma1988neural} removes nodes which have little effect on the overall network output and nodes that are duplicated by other nodes. The author also discusses removing entire layers, if they are found to be redundant \citep{waugh1994dynamic}. Skeletonization is based on the same idea of the network's sensitivity to node removal and proposes to remove nodes from the network based on their relevance during training \citep{skeletonization}.

Optimal brain damage (OBD) uses second-derivative information to automatically delete parameters based on the "saliency" of each paramter - reducing the number of parameters by a factor of four and increasing its recognition accuracy slightly on a state-of-the-art network \citep{lecun1990optimal}. Optimal Brain Surgeon (OBS) enhances the OBD algorithm by dropping the assumption that the Hessian matrix of the neural network is diagonal (they report that in most cases, the Hessian is actually strongly non-diagonal), and they report even better results \citep{hassibi1993optimal}. The algorithm was extended again by the same authors \citep{hassibi1994optimal}.

However, methods based on sensitivity measures have the disadvantage that they do not detect correlated elements - such as two nodes that cancel each other out and could be removed without affecting the networks performance \citep{reed1993pruning}. 

\subsubsection{Constructive Dynamic Learning}

In constructive dynamic learning, one starts with a minimal network structure and iteratively adds complexity to the network by adding new nodes or new connections to existing nodes. 

Two algorithms for the dynamic construction of feed-forward neural networks are presented in this section: the cascade-correlation algorithm (Cascor) and the forward thinking algorithm.

Other algorithms have been proposed but, for brevity, will not be included in this paper's analysis (node splitting \citep{wynne1992node}, the tiling algorithm \citep{mezard1989learning}, the upstart algorithm \citep{frean1990upstart}, a procedure for determining the topology for a three layer neural network \citep{wang1994procedure}, and meiosis networks that replace one "overtaxed" node by two nodes \citep{hanson1990meiosis}).

\paragraph{Cascade-correlation Networks}\mbox{}\\
\label{ss:cascor} 

\begin{figure}[h]
  \center 
  \includegraphics[width=0.8\textwidth]{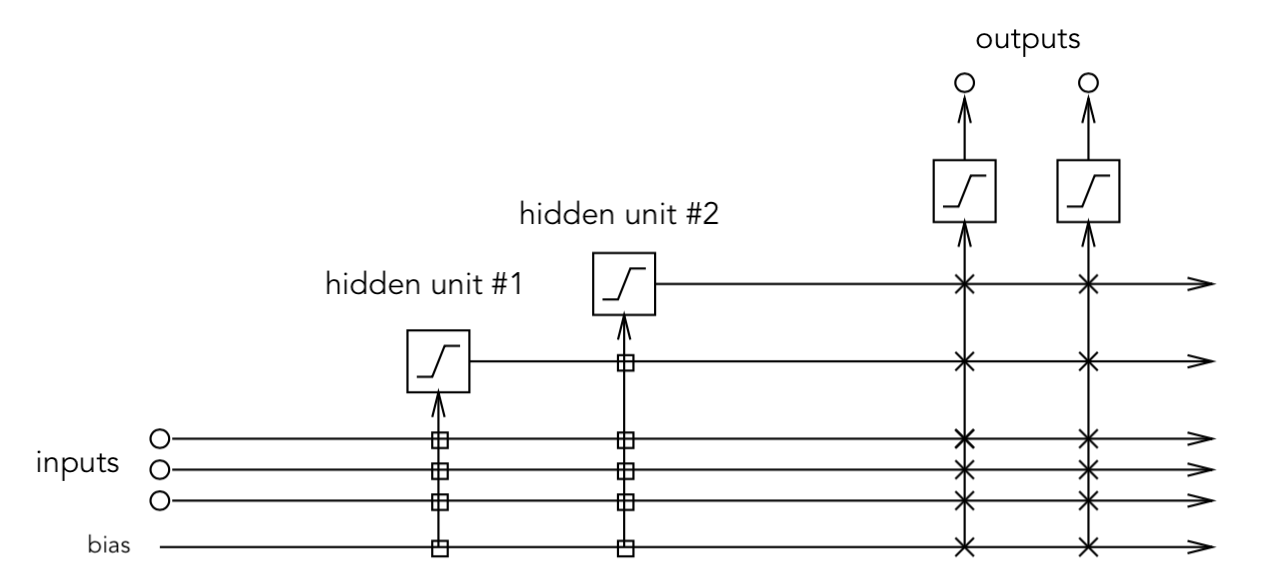}
  \caption{The cascade correlation neural network architecture after adding two hidden units. Squared connections are frozen after training them once, crossed connections are retrained in each training iteration. Figure taken and adapted from \cite{cascor}.}
  \label{fig:cascor}
\end{figure}

The cascade-correlation learning architecture (short: Cascor) was proposed by \cite{cascor}. It is a supervised learning algorithm for neural networks that continuously adds units into the network, trains them one by one and then freezes those unit's input connections. This results in a network that is not layered but has a structure in which all input units are connected to all hidden units and the hidden units have a hierarchical ordering in which the one hidden unit's output is fed into subsequent hidden units as input. When training, Cascor keeps a "pool" of candidate units - possibly using different nonlinear activation functions - and chooses the best candidate unit. Figure \ref{fig:cascor} visualizes this architecture. So-called residual neural networks have been very successful in tasks such as image recognition \citep{he2016deep} through the use of similar skip connections. Cascor takes the idea of skip connections and applies it to include network connections from the input to every hidden node in the network.

Cascor aims to solve two main problems that are found in the widely used backpropagation algorithm: the \textit{step-size problem}, and the \textit{moving target problem}. 

The step size problem occurs in gradient descent optimization methods because it is not clear how big the step in each parameter update should be. If the step size is too small, the network takes too long to converge to a local minimum, if it is too large, the learning algorithm will jump past local minima and possibly not converge to a good solution at all. Among the most successful ways of dealing with this step size problem are higher-order methods, which compute second derivatives in order to get a good estimate of what the step size should be (which is very expensive and often times intractable), or some form of "momentum", which keeps track of earlier steps taken to make an educated guess about how large the step size should be at the current step. 

The moving target problem occurs in most neural networks when all units are trained at the same time and cannot communicate with each other. This leads to all units trying to solve the same learning task - which changes constantly. Fahlman and Lebiere propose an interesting manifestation of the moving target problem which they call the "herd effect". Given two sub-tasks, A and B, that must be performed by the hidden units in a network, each unit has to decide independently which of the two problems it will tackle. If task A generates a larger or more coherent error signal than task B, the hidden units will tend to concentrate on A and ignore B. Once A is solved, the units will then see B as a remaining source of error. Units will move towards task B and, in turn, problem A reappears. Cascor aims to solve this moving target problem by only training one hidden unit at a time. Other approaches, such as the forward thinking formulation, are less restricted and allow the training of one entire layer of units at a time \citep{forwardthinking}.

In their original paper, Fahlman and Lebiere reported good benchmark results on the two-spirals problem and the n-input parity problem. The main advantages over networks using backpropagation were faster training (though this might also be attributed to the use of the Quickprop learning algorithm), deeper networks without problems of vanishing gradients, possibility of incremental learning and, in the n-input parity problem, fewer hidden units in total.

In the literature, Cascor has been criticized for poor performance on regression tasks due to an overcompensation of errors which comes from training on the error correlation rather than on the error signal directly (\citep{littmann1992cascade}, \citep{prechelt1997investigation}). Cascor has also been criticized for the use of its cascading structure rather than adding each hidden unit into the same hidden layer.

\cite{littmann1992cascade} present a different version of Cascor that is based on error minimization rather than error correlation maximization, called Caser. They also present another modified version of Cascor, called Casqef, which is trained on error minimization and uses additional non-linear functions on the output of cascaded units. Caser doesn't do any better than Cascor, while Casqef outperforms Cascor in more complicated tasks - likely because of the additional nonlinearities introduced by the nonlinear functions on the cascaded units.

\cite{littmann1993generalization} show that Cascor is favorable for "extracting information from small data sets without running the risk of overfitting" when compared with shallow broad architectures that contain the same number of nodes. However, this comparison does not take into account deep layered architectures that are popular in today's deep learning landscape.

\cite{sjogaard1991conceptual} suggests that the cascading of hidden units has no advantage over the same algorithm adding each unit into the same hidden layer.

\cite{prechelt1997investigation} finds that Cascor's cascading structure is sometimes better and sometimes worse than adding all the units into one single hidden layer - while in most cases it doesn't make a significant difference. They also find that training on covariance is more suitable for classification tasks while training on error minimization is more suitable for regression tasks. 

\cite{yang1991experiments} find that in their experiments, Cascor learns 1-2 orders of magnitude faster than a network trained with backpropagation, results in substantially smaller networks and only a minor degradation of accuracy on the test data. They also find that Cascor has a large number of design parameters that need to be set, which is usually done through exploratory runs which, in turn, translates into increased computational costs. According to the authors, this might be worth it "if the goal is to find relatively small networks that perform the task well" but "it can be impractical in situations where fast learning is the primary goal". 

Most of the literature available for Cascor is over 20 years old. Cascor seems to not have been actively investigated in recent years. Through email correspondence with the original paper's author, Scott E. Fahlman at CMU, and his PhD student Dean Alderucci, I was made aware of the fact that research on Cascor has been inactive for over twenty years. However, Dean is currently working on establishing mathematical proofs involving how Cascor operates, and adapting the recurrent version of Cascor tosentence classifiers and possibly language modeling. With my experiments, I am starting a preliminary investigation into whether Cascor is still a promising learning algorithm after two decades.

\paragraph{Forward Thinking}\mbox{}\\

In 2017, \cite{forwardthinking} proposed a general framework for a greedy training of neural networks one layer at a time, which they call "forward thinking". They give a general mathematical description of the forward thinking framework, in which one layer is added at a time, then trained on the desired output and finally added into the network while freezing the layer's input weights and discarding its output weights. There are no skip connections, as in Cascor. The goal is to make the data "more separable", i.e. better behaved after each layer. 

In their experiments, \cite{forwardthinking} used a fully-connected neural network with four hidden layers to compare training using forward thinking against traditional backpropagation. They report similar test accuracy and higher training accuracy with the forward thinking network - which hints at overfitting, thus more needs to be done for regularization in the forward thinking framework. However, forward thinking was significantly faster. Training with forward thinking was about 30\% faster than backpropagation - even though they used libraries which were optimized for backpropagation. They also showed that a convolutional network trained with forward thinking outperformed a network trained with backpropagation in training accuracy, testing accuracy while each epoch took about 50\% less time. In fact, the CNN trained using forward thinking achieves near state-of-the-art performance after being trained for only 90 minutes on a single desktop machine.

Both Cascor and forward thinking construct neural networks in a greedy way, layer by layer. However, forward thinking trains layers instead of individual units and while Cascor uses old data to train new units, forward thinking uses new, synthetic data to train a new layer.

\subsubsection{Combined Destructive and Constructive Dynamic Learning}

As mentioned before, it is also possible to combine the destructive and constructive approach to dynamic learning. I was not able to find any algorithms that fit into this area, aside from \cite{waugh1994dynamic}, who proposed a modification to Cascor which also prunes the network. 

\subsection{Summary}

Many current state-of-the-art machine learning solutions rely on deep neural networks with architectures much larger than necessary in order to solve the task at hand. Through early stopping, dropout and other regularization techniques, these overly large networks are prevented from overfitting on the data. Finding a way to efficiently automate the architecture design of neural networks could lead to better network architectures than previously used. In the beginning of this section, I have presented some evidence for neural network architectures that have been designed by algorithms and outperform manually designed architectures. 

Automated architecture design algorithms might be the next step in deep learning. As deep neural networks continue to increase in complexity, we may have to leverage neural architecture search algorithms and dynamic learning algorithms to design deep lerning systems that continue to push the boundary of what is possible with machine learning.

Several algorithms have been proposed to dynamically and automatically choose a neural network's architecture. This thesis aims to give an overview of the most popular of these techniques and to present empirical results, comparing these techniques on different benchmark problems. Furthermore, in the following sections, I will also be introducing new algorithms, based on existing algorithms.

  \section{Empirical Findings}
\label{ss:investigation}


\subsection{Outline of the Investigation}

So far, this thesis has demonstrated the relevance of deep neural networks in today's machine learning research and shown that deep neural networks are more powerful in representing and learning complex functions than shallow neural networks. I have also outlined downsides to using such deep architectures; the trial and error approach to designing a neural network's architecture and the computational inefficiency of oversized architectures that is found in many modern deep learning solutions. 

In a preliminary literature review of possible solutions to combat the computational inefficiencies of deep learning in a more automated, dynamic way, I presented a few algorithms and techniques which aim to automate the design of deep neural networks. I introduced different categories of such techniques; search algorithms, constructive algorithms, destructive algorithms (including regularization techniques), and mixed constructive and destructive algorithms.

I will furthermore empirically investigate a chosen subset of the presented techniques and compare them in terms of final performance, computational requirements, complexity of the resulting model and level of automation. The results of this empirical study may give a comparison of these techniques' merit and guide future research into promising directions. The empirical study may also result in hypotheses about when to use the different algorithms that will require further study to verify.

As the scope of this thesis is limited, the results that will be presented hereby will not be sufficient to confirm or reject any hypotheses about the viability of different approaches to automated architecture design. The experiments presented in this program will act only as a first step of the investigation into which algorithms are worthy of closer inspection and which approaches may be suited for different learning tasks.

\subsubsection{Investigated Techniques for Automated Architecture Design}

The investigated techniques for automated architecture design have been introduced in Section \ref{ss:aut-arch-des}. This section outlines the techniques that will be investigated in more detail in an experimental comparison. 

As search-based techniques for neural network architecture optimization, I will investigate random search and evolving neural networks.

Furthermore, I am running experiments on the cascade-correlation learning algorithm and forward thinking neural networks as algorithms for the dynamical building of neural networks during training. In these algorithms, only one network is considered but each layer is chosen from a set of possible layers from which the best one is chosen.

I will not start an empirical investigation of destructive dynamic learning algorithm. I do not consider any of the introduced destructive dynamic learning algorithms as \emph{automated}. Neither regularization nor pruning existing networks contribute to the automation of neural network architecture design. They are valuable techniques that can play a role in the design of neural networks, in order to reduce the model's complexity and/or improve the network's peformance. However, as they are not \emph{automated} algorithms, I will not be considering them in my empirical investigation.

I furthermore declare the technique of manual search - the design of neural networks through trial and error - as the baseline for this experiment. 

The following list shows all techniques that are to be investigated empirically:
\begin{itemize}
  \item Manual search (baseline)
  \item Random search
  \item Evolutionary search
  \item Cascade-correlation networks
  \item Forward thinking networks
\end{itemize}

\subsubsection{Benchmark Learning Task}

In order to compare different automated learning algorithms, a set of learning tasks need to be decided on which each architecture will be trained, in order to assess their performance. Due to the limited scope of this research project, I will limit myself to the MNIST digit recognition dataset. 

MNIST is the most widely used dataset for digit recognition in machine learning, maintained by \cite{mnist}. The dataset contains handwritten digits that are size-normalized and centered in an image of size 28x28 with pixel values ranging from 0 to 255. The dataset contains 60,000 training and 10,000 testing examples. Benchmark results reported using different machine learning models are listed on the website \href{http://yann.lecun.com/exdb/mnist/}{here}. The resulting function is
$$
  f_{mnist}: \{0,..,255\}^{784} \mapsto \{0,..,9\}
$$
where 
$$
  f_{mnist}(x) = i \text{ iff $x$ shows the digit $i$}
$$

The MNIST dataset is divided into a training set and a testing set. I further divide the training set into a training set and a validation set. The validation set consists of 20\% of the training data. From this point onwards, I will be referring to the training set as the 80\% of the original training set that I am using to train the algorithms and the validation set as the 20\% of the original training set that I am using for a performance metric during training. The testing set will not be used until the final model architecture is decided on. All model decisions (e.g. early stopping) will be based on the network's performance on the validation and training data - not the testing data.

\subsubsection{Evaluation Metrics}

The goal of neural network design was discussed in Section \ref{ss:nn-design}. Based on this, the following list of metrics shows how the different algorithms will be compared and assessed:
\begin{itemize}
  \item Model performance: assessed by accuracy on the unseen testing data.
  \item Computational requirements: assessed by the duration of training (subject to adjustments, due to code optimization and computational power difference between machines running the experiment).
  \item Model complexity: assessed by the number of connections in the resulting network.
  \item Level of automation: assessed by the number of parameters that require optimization.
\end{itemize}

\subsubsection{Implementation Details}
\label{ss:implementation-details}

I wrote the code for the experiments entirely by myself, unless otherwise specified. All my implementations were done in Keras, a deep learning framework in Python, using Tensorflow as a backend. Implementing everything with the same framework makes it easier to compare metrics such as training time easier. 

All experiments were either run on my personal computer's CPU or on a GPU cloud computing platform called Google Colab. Google Colab offers free GPU power for research purposes. More specifically, for the experiments I had access to a Tesla K80 GPU with 2496 CUDA cores, and 12GB of GDDR5 VRAM. My personal computer uses a 3.5 GHz Intel Core i7 CPU with 16 GB of memory.

Some terminology is used without being formally defined. The most important of these terms are defined in the appendix, such as activation functions, loss functions and optimization algorithms that are used in the experiments.

\subsection{Search Algorithms}

The most natural way to find a good neural network architecture is to \emph{search} for it. While the training of a neural network is an optimization problem itself, we can also view the search for an optimal (or simply, a good) neural network architecture as an optimization problem. Within the space of all neural network architectures (here only feedforward architectures), we want to find the architecture yielding the best performance (for example, the lowest validation error). 

The obvious disadvantage is that searching is very expensive. A normal search consists of different stages. First, we have to define the search space, i.e. all neural network architectures that we will be considering in our search. Second, we will search through this space of architectures, assessing the performance of each neural networks by training it until some stopping criterion (depending on the time available, one often does not train the networks until convergence). Third, one evaluates the search results and the performance of each architecture. Now, one can fully train some (or simply one) of the best candidates. Alternatively, we can use the information from the search results to restrict our search space and re-run the search on this new, restricted search space.

It is important to note that this is not an ideal approach. Ideally, one would train each network architecture to convergence (even multiple times, to get a more reliable performance metric) and then choose the best architecture. However, in order to save time, we only train each network for a few epochs and assess its performance based on that. There are other performance estimation techniques \citep{elsken2018neural}, however in these experiments I will train networks for a few epochs and assess their performance based on the resulting accuracy on the testing data. However, as a result of this performance estimation, the search results may be biased to prefer network architectures that perform well in the first few epochs.

\subsubsection{Manual Search}
\label{ss:manual-search}

One of the most widely used approaches by researchers and students is manual search \citep{elsken2018neural}. I also found the names \emph{Grad Student Descent} or \emph{Babysitting} for it. This approach is 100\% manual and based on trial and error, as well as personal experience. One iterates through different neural network setups until one runs out of time or reaches some pre-defined stopping criterion. 

I am also including a research step: researching previously used network architectures that worked well on the learning task (or on similar learning tasks). I found an example MLP architecture on the MNIST dataset in the code of the Keras deep learning framework. They used a feedforward neural network with two hidden layers of 512 units each, using the rectified linear units (relu) activation function and a dropout (with the probability of dropping out being $p=0.2$) after each hidden layer. The output layer uses the softmax activation function (see Appendix \ref{ss:activation}). The network is optimized using the Root Mean Square Propagation algorithm (RMSProp, see Appendix \ref{ss:rmsprop}), with the categorical crossentropy as a loss function (see Appendix \ref{ss:loss-functions}). They report a test accuracy of 98.40\% after 20 epochs \citep{kerasmnist}.

\begin{figure}[h]
  \center
  \includegraphics[width=\textwidth]{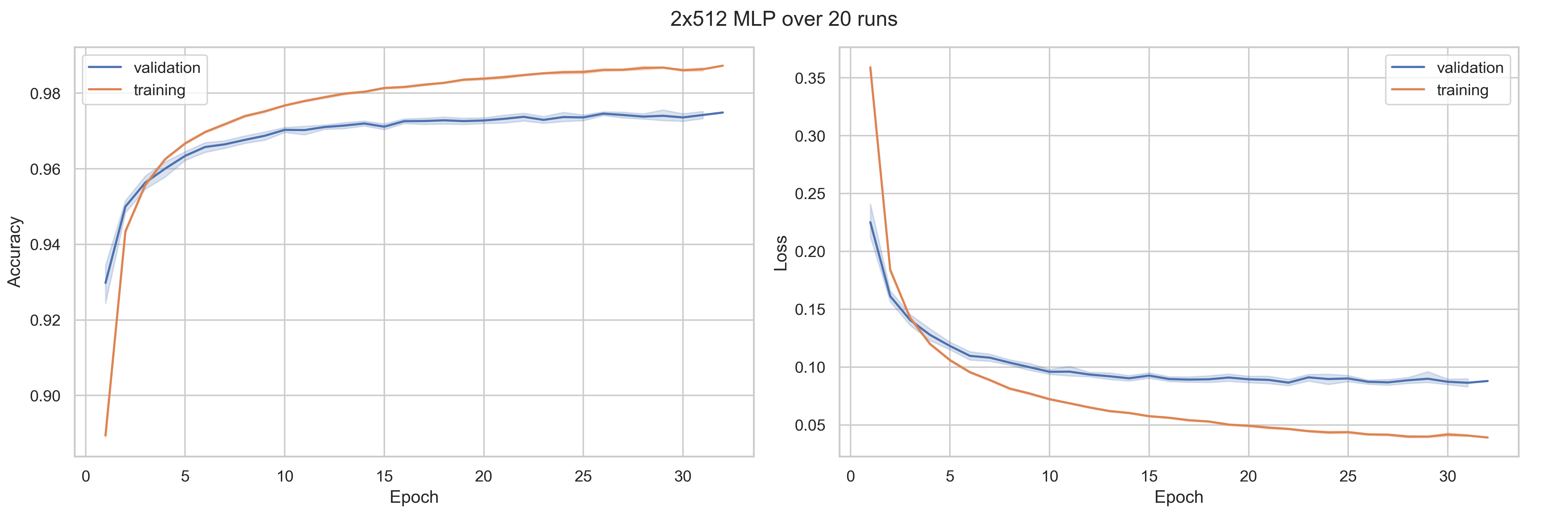}
  \caption{Performance of the neural network found using manual search. Two hidden layers of 512 units each, using the tanh activation function in the hidden units and softmax in the output layer. Trained using RMSProp. Values averaged over 20 training runs.}
  \label{fig:manual-search}
\end{figure}

For this thesis, I do not consider regularization techniques such as dropout, hence I am training a similar network architecture without using dropout. I trained a 2x512 neural network using relu which didn't perform very well so I used the tanh activation function instead - classic manual search, trying different architectures manually. The final network's performance over the training epochs is shown in Figure \ref{fig:manual-search}.

The network's average accuracy on the testing set is 97.3\% with a standard deviation of 0.15\%. The training is stopped after an average of 23 epochs (standard deviation 5.5), after the validation accuracy has not improved for five epochs in a row. Since I am not using dropout (which is likely to improve performance), this result is in agreement with the results reported by \cite{kerasmnist}.

\subsubsection{Random Search}
\label{ss:random-search}

As mentioned in Section \ref{sss:nonadaptive-search}, random search is a good non-adaptive search algorithm \citep{bergstra2012random}. For this thesis, I implemented a random search algorithm to find a good network architecture (not optimizing hyperparameters for the learning algorithm). I start by defining the search space; it consists of:

\begin{itemize}
  \item Topology: how many hidden units per layer and how many layers in total. The number of hidden units per layer $h$ is specified to be $100 \leq h \leq 1000$ (for simplicity, using only multiples of 50) and the number of hidden layers $l$ is specified to be $1 \leq l \leq 10$. 
  \item Activation function: either the relu or tanh function in the hidden layers. The activation function on the output units is fixed to be softmax.
  \item Optimization algorithm: either stochastic gradient descent (SGD) (fixed learning rate, weight decay, using momentum, see Appendix \ref{s:nn-optimization-algorithms}) or RMSProp.
\end{itemize}

Including the topology and activation function in the search space is necessary, as the goal is to search for a good network architecture. I chose not to optimize other hyperparameters, as the focus is to find a good network architecture. However, I did include the choice of optimization algorithm (SGD or RMSProp) to ensure that the optimization algorithm cannot be blamed for bad performance of the networks. As shown in the experiments, RMSProp almost always outperformed SGD. Though I could have only used RMSProp as an optimization algorithm, I chose to leave the optimizer in the search space in order to assess how well the search algorithms performs with "unnecessary" parameters in the search space (unnecessary because RMSProp is better than SGD in all relevant cases, as shown later). 

The program will randomly sample 100 configurations from the search space. Each of the sampled networks will be trained on the training data for five epochs and the performance will be assessed on the training set and the testing set. In order to reduce the noise in the experiment, each network will be trained three times, with different initial weights. All networks are trained using categorical crossentropy loss (see Appendix \ref{ss:loss-functions} with a batch size of 128 (see Appendix \ref{s:nn-optimization-algorithms}).

Table \ref{tbl:best-rand-search-results} shows the ten best results of the experiment. It becomes immediately obvious that RMSProp is a better fit as training algorithm than SGD, as mentioned above. Tanh seems to outperform relu as an activation function in most cases. However, deep and narrow (few hidden units in each layer, with more than five layers) seem to perform better when trained using the relu activation function. 

\begin{table}[h]
\begin{center}
\begin{tabular}{| c | c | c | c | c | c |}
  \hline
  Time & Test acc & Train acc & Activation & Layers & Optimizer\\
  \hline
  7.76s & 96.41\% & 96.11\% & relu & 9 x 100 & RMSProp\\
  6.20s & 96.00\% & 95.78\% & tanh & 3 x 800 & RMSProp\\
  5.19s & 95.85\% & 95.86\% & tanh & 2 x 700 & RMSProp\\
  5.44s & 95.68\% & 95.66\% & tanh & 3 x 550 & RMSProp\\
  5.63s & 95.56\% & 95.85\% & tanh & 2 x 800 & RMSProp\\
  6.20s & 95.51\% & 95.91\% & relu & 6 x 150 & RMSProp\\
  5.00s & 95.42\% & 95.66\% & tanh & 2 x 550 & RMSProp\\
  6.16s & 95.30\% & 95.23\% & tanh & 4 x 600 & RMSProp\\
  5.18s & 95.18\% & 95.17\% & tanh & 3 x 350 & RMSProp\\
  5.61s & 95.06\% & 94.72\% & tanh & 4 x 300 & RMSProp\\
  \hline
\end{tabular}
\end{center}
\caption{Ten best-performing network setups from random search results. All networks trained using categorical cross entropy with softmax in the output layer. Values are averaged over three training runs. Each network was trained for three epochs.}
\label{tbl:best-rand-search-results}
\end{table}

A similar architecture to the two layer architecture from Section \ref{ss:manual-search} shows up in rank 3, showing that manual search yielded a network setup performing (almost) as well as the best network setup found through the random search experiment. However, note that these are only preliminary results - the networks were only trained for three epochs, not until convergence.

It is important to note that the experiment was by far not exhaustive: many hyperparameters were not considered in the random search and the parameters that were considered did not cover all possible choices. This is a comparative study, hence the results of the random search algorithm are only meaningful in comparison to other automated architecture design algorithms.

I continued by training the ten best-performing candidates (based on the averaged accuracy on the validation set) found through the random search experiment until convergence (using early stopping, I stopped training the network once the accuracy on the validation set did not increase for five epochs in a row), I obtain the results shown in Table \ref{tbl:full-rand-search-trained}, sorted by their final performance on the test data. 

\begin{table}[h]
  \begin{center}
  \begin{tabular}{| c | c | c | c | c | c |}
    \hline
    Epochs & Train acc & Test acc & Layers & Activation & Time\\
    \hline
    18 $\pm$ 5 & 98.3\% $\pm$ \ 0.2\% & 97.3\% $\pm$ \ 0.2\% & 2 x 800 & tanh & 31.2s $\pm$ \ 8.1s \\
    24 $\pm$ 5 & 98.5\% $\pm$ \ 0.2\% & 97.2\% $\pm$ \ 0.2\% & 2 x 550 & tanh & 37.8s $\pm$ \ 8.0s \\
    19 $\pm$ 5 & 98.3\% $\pm$ \ 0.2\% & 97.1\% $\pm$ \ 0.5\% & 2 x 700 & tanh & 30.6s $\pm$ \ 8.0s \\
    22 $\pm$ 5 & 98.2\% $\pm$ \ 0.2\% & 97.0\% $\pm$ \ 0.2\% & 3 x 350 & tanh & 36.9s $\pm$ \ 8.7s \\
    18 $\pm$ 4 & 98.3\% $\pm$ \ 0.2\% & 97.0\% $\pm$ \ 0.2\% & 3 x 550 & tanh & 31.0s $\pm$ \ 6.3s \\
    18 $\pm$ 5 & 98.1\% $\pm$ \ 0.3\% & 96.9\% $\pm$ \ 0.3\% & 3 x 800 & tanh & 34.8s $\pm$ 10.5s \\
    26 $\pm$ 5 & 98.1\% $\pm$ \ 0.2\% & 96.8\% $\pm$ \ 0.1\% & 4 x 300 & tanh & 44.8s $\pm$ \ 8.1s \\
    17 $\pm$ 5 & 97.9\% $\pm$ \ 0.3\% & 96.7\% $\pm$ \ 0.5\% & 9 x 100 & relu & 38.5s $\pm$ 12.9s \\
    20 $\pm$ 6 & 97.9\% $\pm$ \ 0.3\% & 96.7\% $\pm$ \ 0.3\% & 4 x 600 & tanh & 38.0s $\pm$ 11.6s \\
    13 $\pm$ 5 & 71.8\% $\pm$ 42.5\% & 70.6\% $\pm$ 41.7\% & 6 x 150 & relu & 26.2s $\pm$ 11.4s \\
    \hline
  \end{tabular}
  \end{center}
  \caption{Best-performing network architectures from random search, sorted by final accuracy on the testing data. The table shows average values and their standard deviations over ten training runs for each network architecture.}
  \label{tbl:full-rand-search-trained}
\end{table}

The results show that the networks using the tanh activation function mostly outperform those using the relu activation function. The best-performing networks are those using two hidden layers, as the one that was trained through manual search. The final performance of the best networks found through random search can be considered equal to the network found through random search.

\subsubsection{Evolutionary Search}
\label{ss:evolutionary-search}

As an adaptive search algorithm, I implemented an evolving artificial neural network which is basically an evolutionary search algorithm applied to neural network architectures, since I am not evolving the connection weights of the network. Evolutionary search algorithms applied to neural networks are also called neuroevolution algorithms. The parameter space is the same as for random search, see Section \ref{ss:random-search}. 

There are several parameters that adjust the evolutionary search algorithm's performance. The parameters that can be adjusted in my implementation are:

\begin{itemize}
  \item Population size: number of network architectures that are assessed in each search iteration.
  \item Mutation chance: the probability of a random mutation taking place (after breeding).
  \item Retain rate: how many of the fittest parents should be selected for the next generation.
  \item Random selection rate: how many parents should be randomly selected (regardless of fitness, after retaining the fittest parents).
\end{itemize}

The listing in Figure \ref{lst:eann} shows a simplified version of the search algorithm. 

\begin{figure}[h]
\begin{lstlisting}[language=Python]
def evolving_ann():
  population = Population(parameter_space, population_size)
  while not stopping_criterion:
    population.compute_fitness_values()
    parents = population.fittest(k)
    parents += population.random(r)
    children = parents.randomly_breed()
    children.randomly_mutate()
    population = parents + children
  return population
\end{lstlisting}
\caption{Simplified pseudo code for the implementation of evolving artificial neural networks}
\label{lst:eann}
\end{figure}

In my implementation, I set the population size to 50, the mutation chance to 10\%, the retain rate to 40\% and the random selection rate to 10\%. These values for the algorithm's parameters were taken from \cite{eannimplementation} and adjusted. The fitness is just the accuracy of the network on the testing set after training for three epochs. As was done in random search, each network is trained three times. The average test accuracy after three epochs is taken as the network's fitness.

In order to make the random search and the evolutionary search experiments comparable, they are both testing the same number of networks. In random search, I picked 200 networks at random. In this evolutionary search algorithm, I stopped the search once 200 networks have been trained. This happened after seven iterations in the evolutionary search.

I ran the algorithm twice, once allowing for duplicate network architectures in the population and once removing these duplicates. 

\paragraph{With duplicates}\mbox{}\\

Without removing duplicate configurations, the search algorithm converges to only six different configurations, shown in Table \ref{tbl:eann-dupe}. The table shows these six configurations.

It is important to note that by allowing duplicate neural network configurations, the algorithm is training multiple instances for each well-performing configuration - hence improving the overall network performance slightly by choosing the best random weight initialization(s).

\begin{table}[h]
\begin{center}
\begin{tabular}{| c | c | c | c | c | c |}
  \hline
  Layers & Optimizer & Hidden & Fitness\\
  \hline
  3 x 450 & RMS Prop & tanh & 95.95\%\\
  4 x 600 & RMS Prop & tanh & 95.90\%\\
  2 x 450 & RMS Prop & tanh & 95.70\%\\
  3 x 350 & RMS Prop & tanh & 95.59\%\\
  2 x 350 & RMS Prop & tanh & 95.45\%\\
  1 x 500 & RMS Prop & tanh & 94.25\%\\
  \hline
\end{tabular}
\end{center}
\caption{Network architectures from evolutionary search without removing duplicate configurations.}
\label{tbl:eann-dupe}
\end{table}

When fully training these configurations, I get the results shown in Table \ref{tbl:eann-dupe-full}. The best network architectures perform similarly to the best ones found through random search. Notably, all networks use tanh as activation function and RMSProp as optimizer.

\begin{table}[h]
\begin{center}
\begin{tabular}{| c | c | c | c | c | c |}
  \hline
  Epochs & Train acc & Test acc & Layers & Activation & Time\\
  \hline
22 $\pm$ 4 & 98.2\% $\pm$ 0.2\% & 97.2\% $\pm$ 0.1\% & 2 x 350 & tanh & 33.8s $\pm$ 5.5s \\
24 $\pm$ 6 & 98.4\% $\pm$ 0.2\% & 97.2\% $\pm$ 0.2\% & 2 x 450 & tanh & 37.7s $\pm$ 10.2s \\
22 $\pm$ 7 & 98.4\% $\pm$ 0.3\% & 97.0\% $\pm$ 0.1\% & 3 x 450 & tanh & 37.2s $\pm$ 11.3s \\
22 $\pm$ 5 & 98.2\% $\pm$ 0.2\% & 96.9\% $\pm$ 0.2\% & 3 x 350 & tanh & 35.7s $\pm$ 8.1s \\
18 $\pm$ 5 & 97.9\% $\pm$ 0.2\% & 96.8\% $\pm$ 0.2\% & 4 x 600 & tanh & 33.8s $\pm$ 8.7s \\
24 $\pm$ 9 & 96.4\% $\pm$ 0.2\% & 96.0\% $\pm$ 0.2\% & 1 x 500 & tanh & 34.2s $\pm$ 13.0s \\
  \hline
\end{tabular}
\end{center}
\caption{Fully trained networks obtained from evolutionary search without removing duplicate configurations.}
\label{tbl:eann-dupe-full}
\end{table}

\paragraph{Without duplicates}\mbox{}\\

When removing duplicate configurations, there will naturally be more variety in the neural network configurations that will appear in later iterations of the search algorithm. Table \ref{tbl:eann-nodupes} shows the ten best neural network configurations found using the evolutionary search algorithm when removing duplicate architectures. 

\begin{table}[h]
\begin{center}
\begin{tabular}{| c | c | c | c | c |}
  \hline
  Layers & Optimizer & Hidden & Test accuracy\\
  \hline
  9 x 150 & RMSProp & tanh & 96.24\%\\
  2 x 850 & RMSProp & tanh & 96.23\%\\
  2 x 950 & RMSProp & tanh & 96.12\%\\
  3 x 500 & RMSProp & tanh & 95.78\%\\
  9 x 100 & RMSProp & tanh & 95.74\%\\
  4 x 600 & RMSProp & tanh & 95.71\%\\
  4 x 800 & RMSProp & tanh & 95.56\%\\
  4 x 400 & RMSProp & tanh & 95.42\%\\
  9 x 100 & RMSProp & tanh & 95.32\%\\
  4 x 650 & RMSProp & tanh & 95.31\%\\
  \hline
\end{tabular}
\end{center}
\caption{Top ten neural network configurations found using EANNs without duplicate configurations.}
\label{tbl:eann-nodupes}
\end{table}

The results are better than the ones obtained from the evolutionary search with duplicate architectures. This is likely due to the increased variety in network architectures that are considered by the search algorithm. Fully training these networks yields the results in Table \ref{tbl:eann-nodupes-full}.

\begin{table}[h]
\begin{center}
\begin{tabular}{| c | c | c | c | c | c |}
  \hline
  Epochs & Train acc & Test acc & Layers & Act. & time\\
  \hline
20 $\pm$ 6 & 98.3\% $\pm$ 0.3\% & 97.3\% $\pm$ 0.1\% & 2 x 850 & tanh & 33.6s $\pm$ 10.3s \\
18 $\pm$ 5 & 98.2\% $\pm$ 0.2\% & 97.2\% $\pm$ 0.3\% & 2 x 950 & tanh & 31.2s $\pm$ 8.4s \\
19 $\pm$ 5 & 98.3\% $\pm$ 0.2\% & 96.9\% $\pm$ 0.2\% & 3 x 500 & tanh & 32.2s $\pm$ 7.8s \\
25 $\pm$ 7 & 98.2\% $\pm$ 0.3\% & 96.8\% $\pm$ 0.2\% & 4 x 400 & tanh & 43.3s $\pm$ 11.7s \\
20 $\pm$ 6 & 98.0\% $\pm$ 0.2\% & 96.7\% $\pm$ 0.2\% & 4 x 600 & tanh & 37.3s $\pm$ 10.7s \\
21 $\pm$ 7 & 97.9\% $\pm$ 0.2\% & 96.7\% $\pm$ 0.3\% & 4 x 650 & tanh & 41.7s $\pm$ 13.4s \\
20 $\pm$ 5 & 97.7\% $\pm$ 0.2\% & 96.7\% $\pm$ 0.2\% & 4 x 800 & tanh & 42.4s $\pm$ 10.5s \\
27 $\pm$ 5 & 96.5\% $\pm$ 0.3\% & 95.5\% $\pm$ 0.3\% & 9 x 150 & tanh & 62.6s $\pm$ 10.9s \\
24 $\pm$ 7 & 95.8\% $\pm$ 0.4\% & 94.9\% $\pm$ 0.5\% & 9 x 100 & tanh & 54.1s $\pm$ 16.5s \\
  \hline
\end{tabular}
\end{center}
\caption{Top ten neural network configurations found using EANNs without duplicate configurations, fully trained (until validation accuracy hasn't improved for five epochs in a row).}
\label{tbl:eann-nodupes-full}
\end{table}

These results are also very similar to the ones obtained through random search and manual search. The best-performing architectures are using two hidden layers, though here the number of neurons in these hidden layers is larger than previously seen.

The animation in Figure \ref{fig:animated-evolution} shows how the population in this evolutionary search algorithm changes between iterations. The animation demonstrates how the accuracy of the networks in the population increases with each search iteration, with some random fluctuations due to the random mutations that are sometimes disadvantageous. It also shows that RMSProp is quickly adopted as the optimizer mainly used in the iterations and that tanh is adopted as the activation function that is mainly used. The model complexity is shown on the x axis and the animation shows that the evolutionary search converges to results at the lower end of the model complexity scale. This confirms that smaller network architectures are more suited for the learning task at hand than larger architectures. 

\begin{figure}[h]
  \center
  \animategraphics[autoplay,loop,width=0.7\textwidth]{2}{assets/evolution_}{1}{7}
  \caption{Animation of how the population in the evolutionary search algorithm changes between iterations (best viewed in Adobe Acrobat).}
  \label{fig:animated-evolution}
\end{figure}

\subsubsection{Conclusion}

All three search algorithms yield the same final performance, with minor differences. They all find that architectures using two hidden layers seem to work the best and only differ in the width of these hidden layers. Hence, the performance of the three search algorithms can be considered equal. 

The complexity of the resulting model (measured by the number of hidden layers and the width of these layers) is also comparable between the three search algorithms, as they find similar network architectures. To be very exact, evolutionary search (when allowing for duplicates in the population) finds the smallest network architecture (two hidden layers of 350 or 450 neurons each), followed by manual search (two hidden layers of 512 neurons each), then random search (two hidden layers of 800, 550, or 700 neurons each) and final evolutionary search (when removing duplicate architectures from the population) with two hidden layers of 850 or 950 neurons each. However, I do not consider these findings very relevant but consider them to be due to random noise in the experiments - multiple runs of the search algorithms will give more statistically significant results and may come up with a different ordering in the resulting network's complexity, since the difference between the network architectures does not seem very significant in the experiments that I ran.

The level of automation differs significantly between the three algorithms. Manual search is obviously not automated at all. Evolutionary search is automated but still has a lot of hyperparameters that need to be decided (listed in Section \ref{ss:evolutionary-search}). Random search is the most automated algorithm, it merely requires the specification of the search space.

The computational requirements for the different search algorithms are difficult to compare. Technically, my implementation of manual search was very efficient - I only trained two network architectures until reaching the architecture that I reported my findings for. However, in practice, manual search is often an iterative process, in which one tries different architectures and decides on an architecture based on this trial and error. This is difficult, if not impossible, to quantify. 
Comparing the random search and evolutionary search algorithm with respect to computational requirements is not straight-forward either. Their space requirements are similar (assuming an efficient way of storing the population in evolutionary search, which is the case in my implementation). The time requirements of the two algorithms is difficult to compare. Due to the random nature of both algorithms, and because I am only reporting one run for each of the search algorithms, it is not possible to compare the algorithm's time requirements in a meaningful way based on the experiments I conducted. 

\begin{figure}[h]
  \center
  \includegraphics[width=\textwidth]{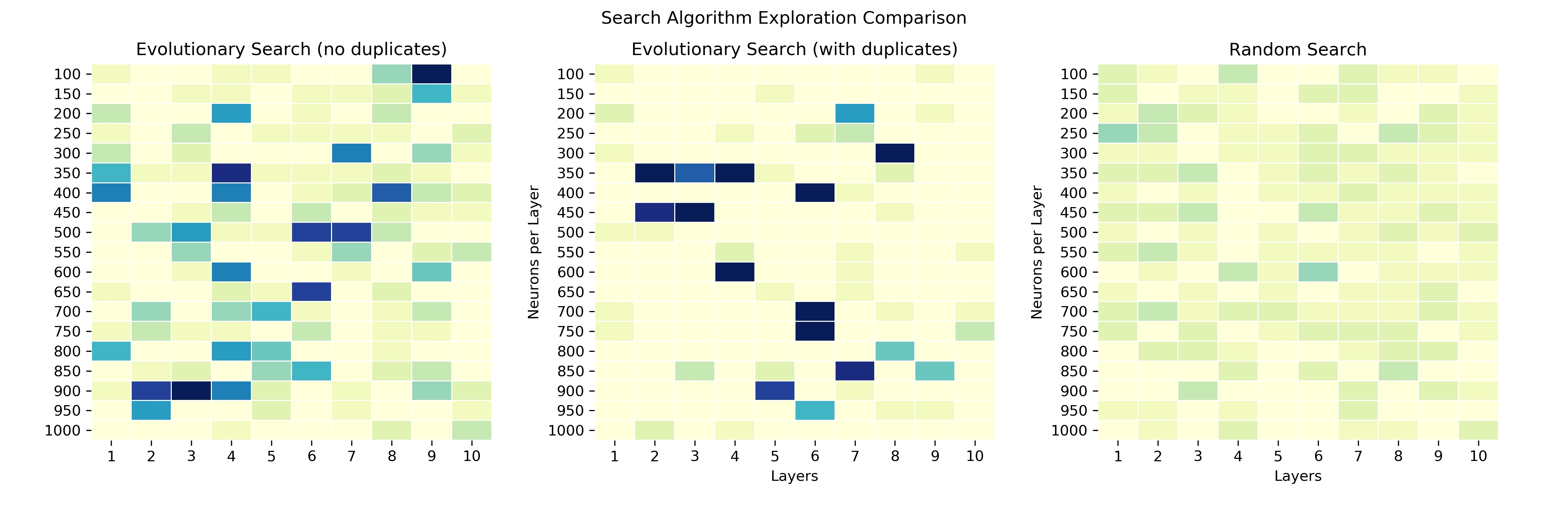}
  \caption{Exploration of the network architecture search space using different search algorithms. Hidden activation function and optimizer are omitted. The color encoding is the same for all three plots.}
  \label{fig:search-exploration}
\end{figure}

A meaningful comparison is the exploration of the search space, i.e. how much of the search space has been explored by the algorithm. Figure \ref{fig:search-exploration} shows how the two version of evolutionary search compare with the random search algorithm. As expected, random search explores the search space very evenly. When removing duplicates in the population, the evolutionary search algorithm explores more of the search space compared to not removing duplicate architectures. When allowing for duplicates, the exploration looks very clustered, indicating that the algorithm mainly stayed in the same areas of the search space. When removing duplicates, the exploration is more spread out, though not as balanced as random search. 

\begin{figure}[h]
  \center
  \includegraphics[width=0.75\textwidth]{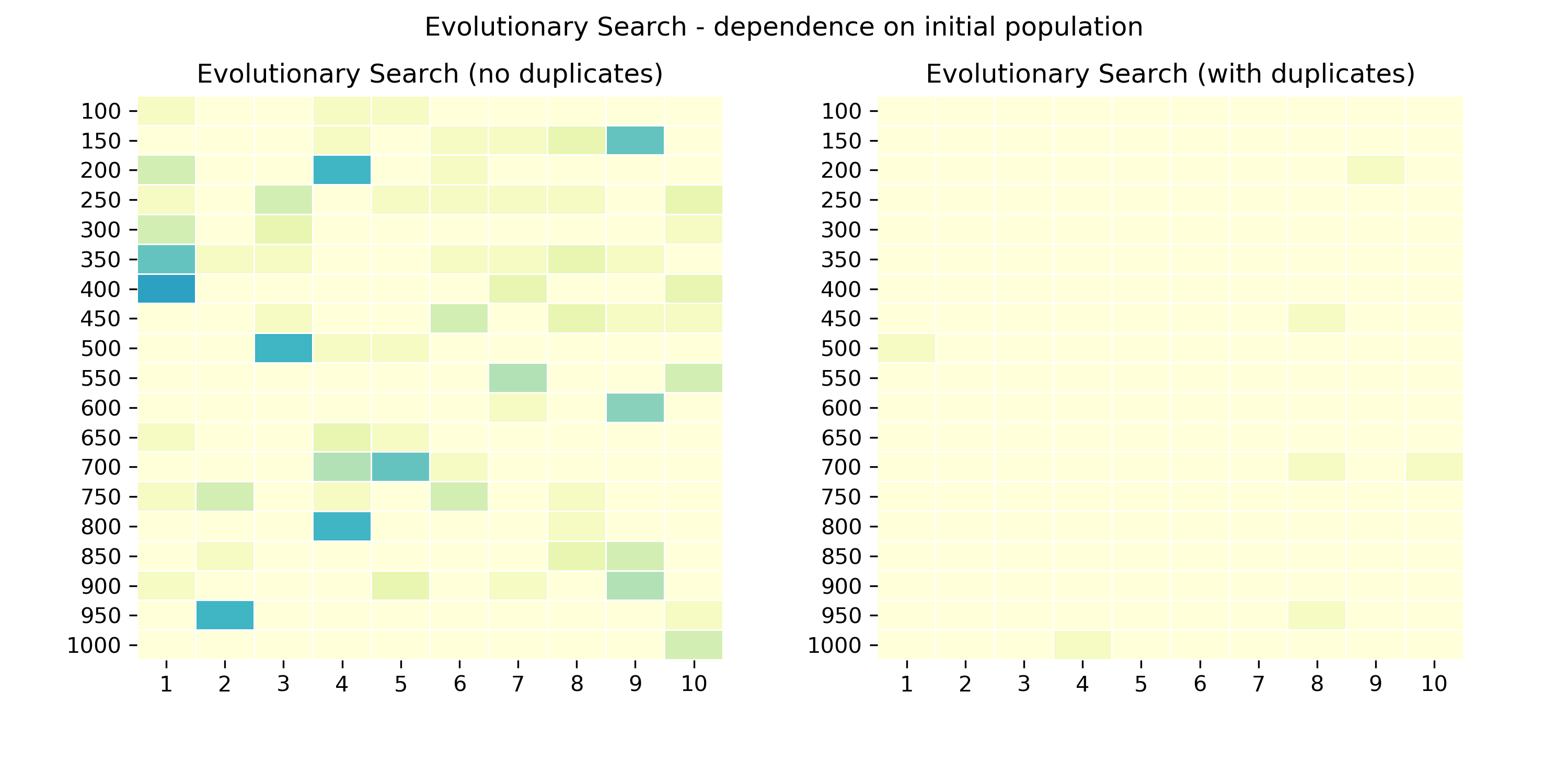}
  \caption{Exploration of the neural architecture search space for evolutionary search (with or without duplicates in the population), when removing all those architectures that were present in the initial population. The lower the activity in the search space, the more the exploration depends on the initial population. Hidden activation function and optimizer are omitted. The color encoding is the same for all three plots.}
  \label{fig:search-dependence-initial}
\end{figure}

The exploration of the evolutionary search algorithm is quite dependent on the initial population. Figure \ref{fig:search-dependence-initial} shows how little the evolutionary search algorithm explores architectures that are not in the initial population. When allowing for duplicates, the algorithm almost exclusively checks the architectures from the initial population - only 2\% of all explored architectures were not in the initial population. When removing duplicates, the algorithm explores significantly more, though the initial population still makes up more than 50\% of all explored network architectures. 

This shows that my evolutionary search algorithm implementation is dependent on the initial population. This opens up the possibility to encode prior knowledge into the evolutionary search. If one knows that a particular kind of network architecture is more likely to perform well than another, this can be represented in the initial population for the search. 

To summarize my findings of different neural network architecture search algorithms, each one of the three search algorithms has its advantages and disadvantages. When the designer of the neural network is knowledgeable and experienced in the design of neural network architectures, or has resources such as previously used networks for the learning tasks available, manual search is a good choice. It is very cheap and highly customizable. When the goal is to automate the architecture design, random search and evolutionary search are more suitable choices. Evolutionary search allows for more customization and the encoding of prior knowledge which may save time during the search. Random search is  good algorithm to explore the entire search space evenly, if the goal is to not overlook any architectures.

\subsection{Constructive Dynamic Learning Algorithm}

In constructive dynamic learning, it is not necessary to define the search space explicitly. However, one can argue that different constructive dynamic learning algorithms have implicit restrictions on the type of network architecture that they consider. The cascade-correlation learning algorithm can only build network architectures that are cascaded in a very particular way. The original forward thinking algorithm requires specification of the exact network architecture, thus not automating the architecture design. This is why I am proposing a new algorithm, based on forward thinking, which also automates the architecture design.

\subsubsection{Cascade-Correlation Networks}
\label{ss:exp-cascor}

The originally proposed Cascor algorithm requires many hyperparameters to be set \citep{yang1991experiments}. It does not specify when to stop training each unit before adding the next one and it does not specify when to stop adding new units altogether. Other papers have also questioned the choice of training on error correlation maximization rather than "standard" error minimization training \citep{littmann1992cascade}. I implemented and ran experiments on several different versions of Cascor, aiming to find a version of Cascor that is suitable to a more modern, higher-dimensional dataset such as MNIST (as opposed to the low dimensional, small datasets used in the original paper by \cite{cascor}). The largest dataset for which I found evidence that Cascor had been trained on is a learning task with 120 inputs and 3,175 samples, and a learning task with 21 inputs and 7,100 samples reported by \cite{littmann1992cascade}. MNIST, the dataset I am using in this thesis, has 784 inputs and 80,000 samples. 

All experiments reported in this section were run on my personal computer, see Section \ref{ss:implementation-details} for details.

The parameters that needed to be decided on for the Cascor algorithm are:
 \begin{itemize}
   \item Activation function
   \item Loss function: the originally proposed error correlation, or error minimization.
   \item When to stop training each unit before adding a new one
   \item When to stop adding new units
 \end{itemize}

\paragraph{Cascor}\mbox{}\\

The originally proposed cascade-correlation learning algorithm was described in Section \ref{ss:cascor}. I implemented the algorithm, as well as the proposed error correlation training. The error correlation loss is described in Appendix \ref{ss:error-correlation}.

\begin{figure}[h]
  \center
  \includegraphics[width=\textwidth]{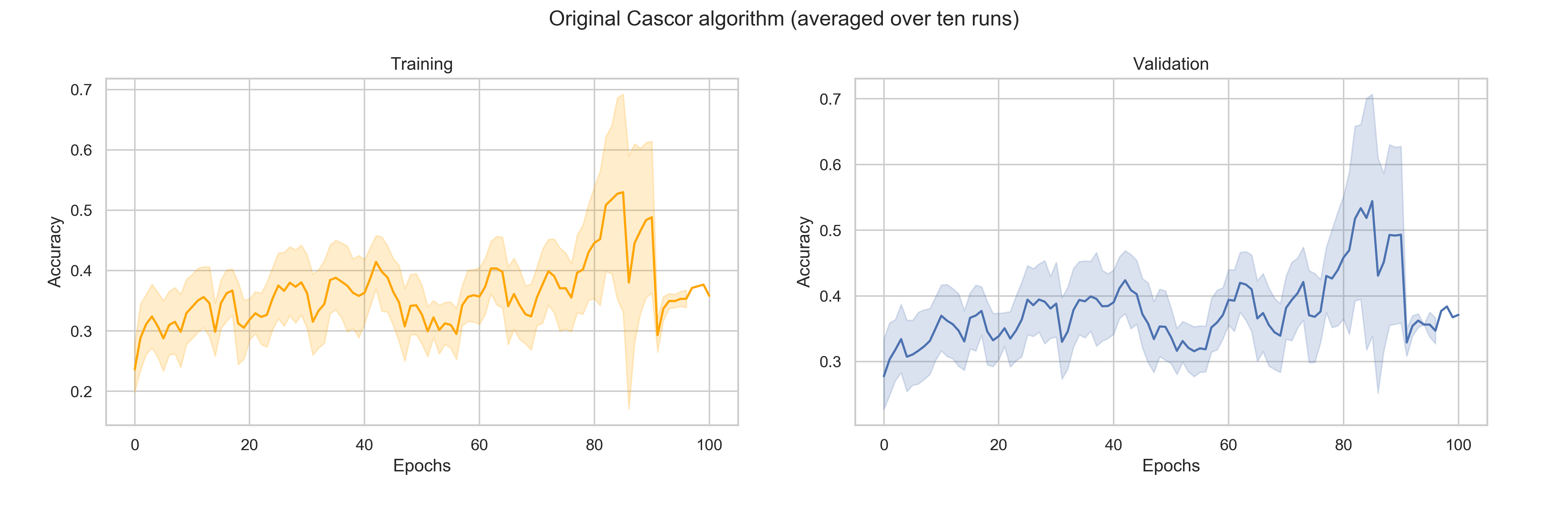}
  \caption{Cascade-correlation learning algorithm, as proposed by \cite{cascor}. The algorithm was run ten times, with a candidate pool of size eight, training each hidden unit in the candidate pool for two epochs and then choosing the one with the highest validation accuracy. This unit is then added into the network and trained until convergence (i.e. until the validation accuracy doesn't improve for three epochs in a row). Results are averaged over the ten runs, with the shaded area representing the 95\% confidence interval.}
  \label{fig:cascor-original}
\end{figure}

The network performs very poorly when trained using the originally proposed error correlation maximization. Training the network several times, it never reached a validation accuracy above 70\%, as shown in Figure \ref{fig:cascor-original}. I have tried different approaches to improve the network's performance but I was not able to report any good findings.

\cite{littmann1992cascade} report that error correlation training is inferior to error minimization training on regression tasks. In classification tasks, it converges faster - though the final performance seems to be the same for both (the authors do not explicitly state so, but it seems to be implied in their conclusion's wording). It may be that the error correlation training overcompensates for errors \cite{prechelt1997investigation} due to the high dimensionality of the dataset, though this requires further investigation.

\paragraph{Caser}\mbox{}\\

\begin{figure}[h]
  \center
  \includegraphics[width=\textwidth]{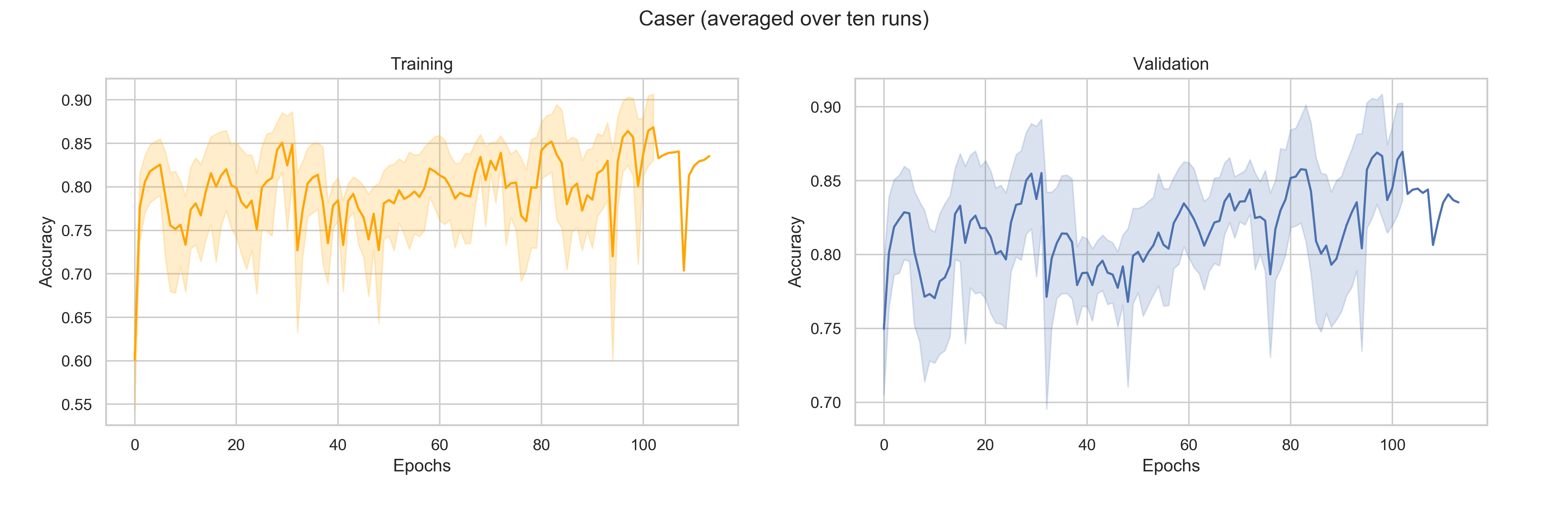}
  \caption{Caser algorithm, as originally proposed by \cite{littmann1992cascade}. Results are averaged over the ten runs, with the shaded area representing the 95\% confidence interval.}
  \label{fig:caser-aggregated}
\end{figure}

The next approach is Caser, as proposed by \cite{littmann1992cascade} - a variation of Cascor in which the network is trained on error minimization. My implementation of the network is using softmax in the output layer, tanh in the hidden units and is trained on the categorical cross entropy loss function. Hidden units are added into the network as described in the original paper. I am using a candidate pool of eight units. Each candidate unit is trained for one epoch after which the candidate unit with the highest accuracy on the validation set is inserted into the network. Once inserted, the unit is trained until convergence using RMSProp (until the testing accuracy stops increasing for more than two epochs in a row) after which the unit's input weights are frozen. The output weight vector is discarded whenever a new unit is added into the network and retrained, similarly to forward thinking. Figure \ref{fig:caser-aggregated} shows the training graphs of this architecture, averaged over ten runs. Overall, this looks much better than the error correlation training in Figure \ref{fig:cascor-original}.

\begin{figure}[h]
  \center
  \includegraphics[width=\textwidth]{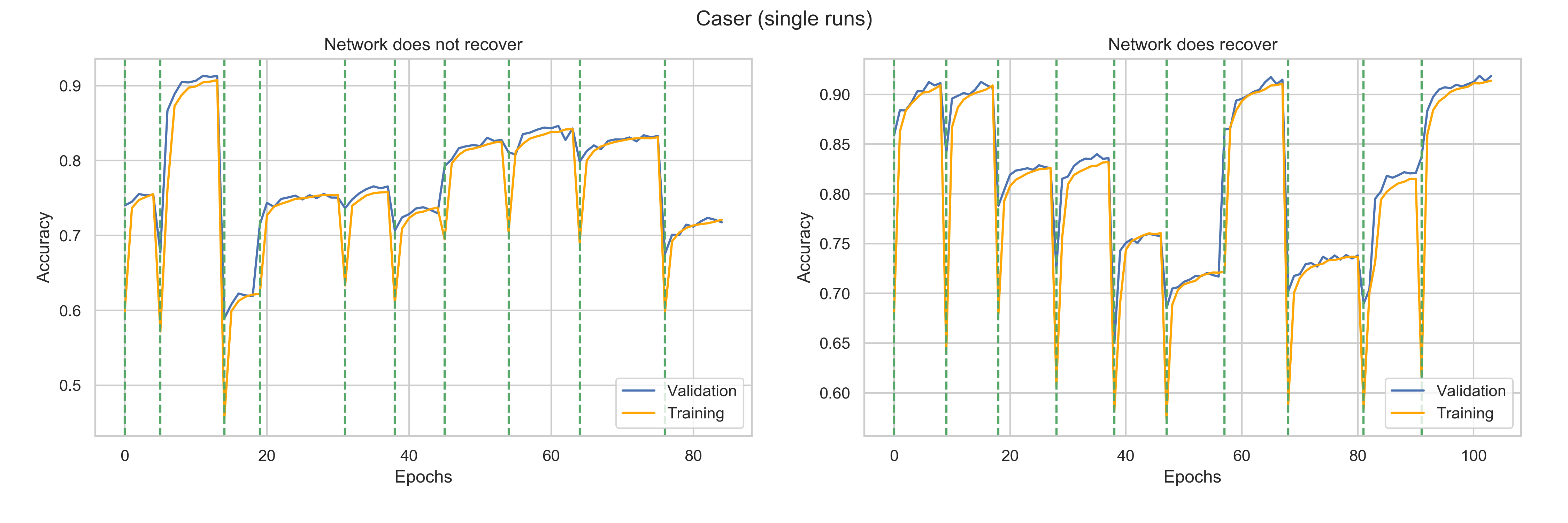} 
  \caption{Unpredictable behavior when adding new units into the Caser network. Left plot shows the Caser network using a candidate pool size of eight, whereas on the right, a candidate pool of size 16 was used. Green dotted lines show the insertion of a new hidden unit into the network.}
  \label{fig:cascor-unpredictable}
\end{figure}

Running this architecture shows some interesting behavior when a new unit is added into the network. Whenever a new hidden unit is added into the network, the network performance changes - sometimes quite drastically. Figure \ref{fig:cascor-unpredictable} shows how unpredictable this turns out in individual training runs. On the left, after adding the second hidden unit, the network accuracy improves to over 90\% but adding a third hidden unit decreases the accuracy down to 60\%, even after training this third unit to convergence. The network never recovers from this performance dip and doesn't reach an accuracy better than 85\% again. This is likely because the output weight vector that the network converged to when training the second hidden unit was discarded and the network will choose a new output weight vector at random (from the pool of eight candidate units). If the candidate pool only contains "bad" weight vectors for the output layer, the network will be stuck in one of these bad local minima. 

In order to remove these sudden (and seemingly uncontrollable) performance jumps, one may increase the candidate pool size, in an attempt to increase the probability of finding a weight vector close to a good local minimum. The right plot in Figure \ref{fig:cascor-unpredictable} shows the performance of a network that uses a candidate pool size of 16 (instead of eight, as the left plot) and shows a large performance decrease after adding the second hidden unit, but recovers to the previous "good" performance with the insertion of the seventh hidden unit. It decreases again with the eighth unit and increases to a new maximum performance with the tenth hidden unit. Luckily, that was the last hidden unit so the final network reaches a good performance. Increasing the candidate pool size is not a deterministic way of finding a better weight vector. A more reliable method is needed to improve Caser's performance.

\paragraph{CaserRe}\mbox{}\\

\begin{figure}[h]
  \center
  \includegraphics[width=\textwidth]{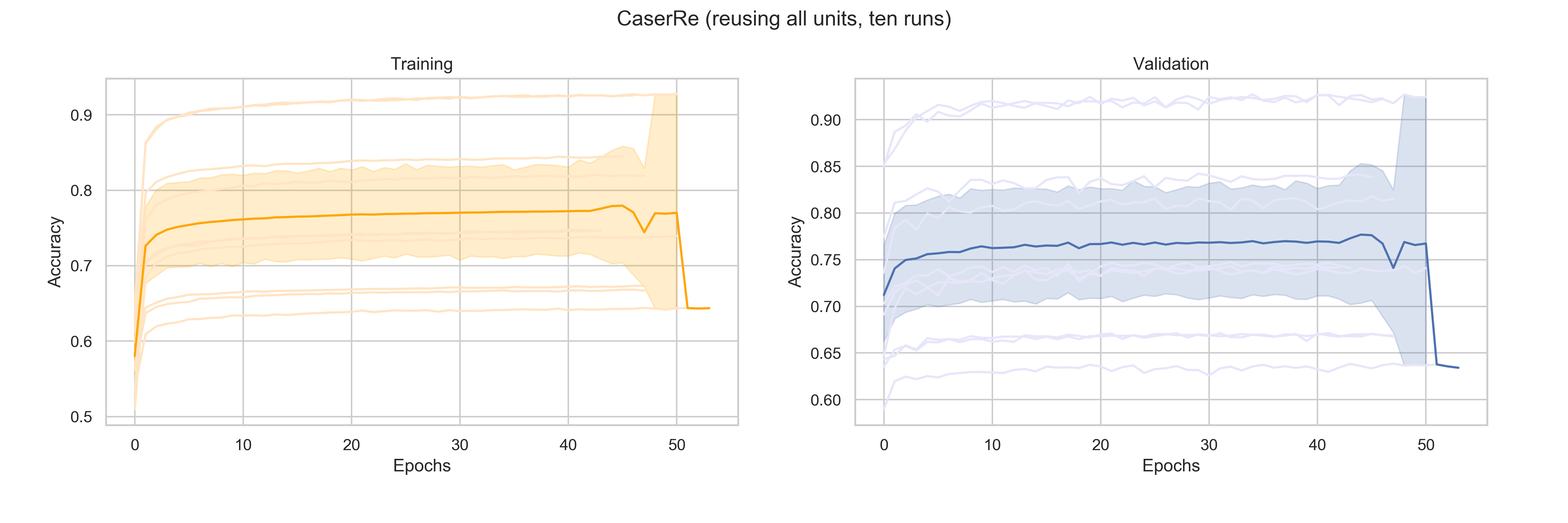}
  \caption{Reusing the output weight for all units in the candidate pool for Caser. Results are averaged over the ten runs, with the shaded area representing the 95\% confidence interval. Lighter colored lines show the single runs.}
  \label{fig:caser-re-all}
\end{figure}

The question of when to stop the training remains, and the random jumps in network performance make it difficult to decide on a stopping criterion. Instead of increasing the candidate pool's size, I initialized the weight vectors for new hidden units close to the local minimum that was found in training the previous hidden unit. As Figure \ref{fig:caser-re-all} shows, this removes performance decreases and yields "smoother" training improvements. I am calling this CaserRe because it is based on Caser and extends it by re-using the output weight vector when a new hidden unit is added into the network.

\begin{figure}[h]
  \center
  \includegraphics[width=\textwidth]{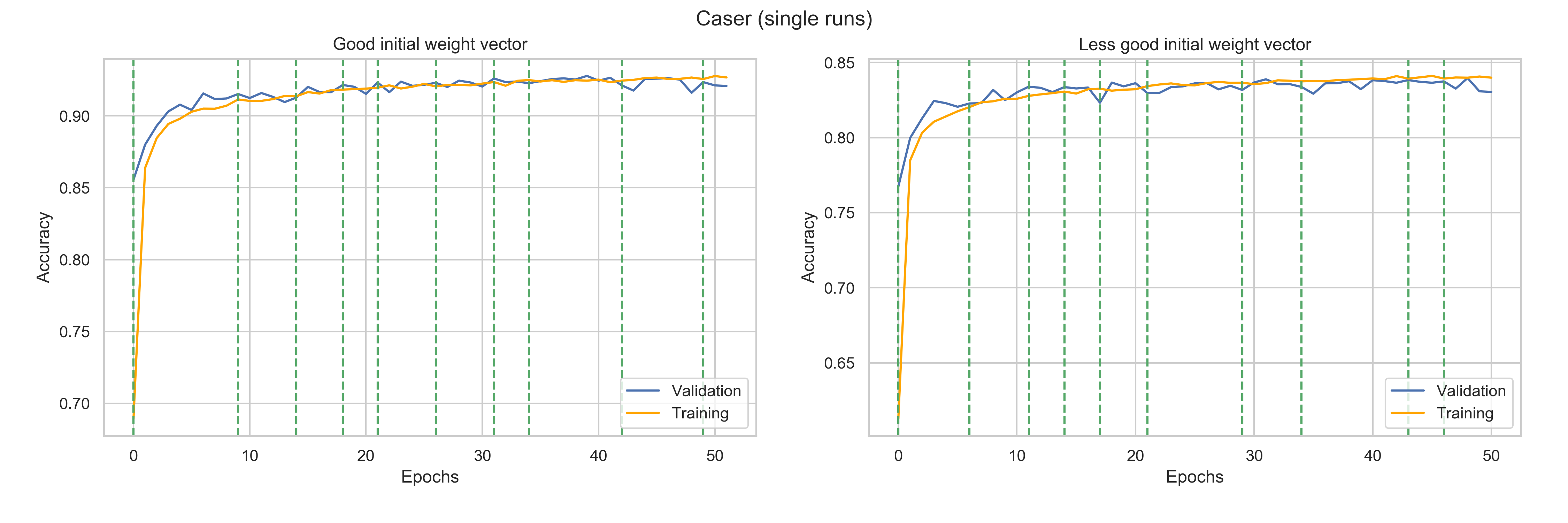}
  \caption{Caser's dependence on the initial weight vector. On the left, the network finds a good initial local minimum whereas on the right, the network finds a worse local minimum and does not improve its performance significantly.}
  \label{fig:caser-dependence}
\end{figure}

However, this makes the network very dependent on the initially found local minimum. By taking the weight vector from the previous hidden unit's training I remove performance dips that would have appeared otherwise - but I also removed performance increases that would otherwise be possible and would help the network jump to a better local minimum. This is shown on individual training runs in Figure \ref{fig:caser-dependence}. If the first hidden unit finds a good local minimum, the overall result will be good, though only slightly improving on the network's performance with one hidden unit. However, if the initial local minimum is not good, the network seems to be stuck.

In order to avoid the pitfalls of a bad weight initialization at the beginning of training, it may help to train the candidate pool of hidden units, choose the best performing hidden unit and, if the performance is not \emph{significantly} worse than it was before adding this hidden unit, the unit should be added as it is. If the performance is significantly worse than before, the unit should be added reusing the previous output weight vector - thus initializing the output weight vector close to the previously found local minimum. This will remove performance dips, while keeping the chance to find better local minima when adding new hidden units. 



\begin{figure}[h]
  \center
  \includegraphics[width=\textwidth]{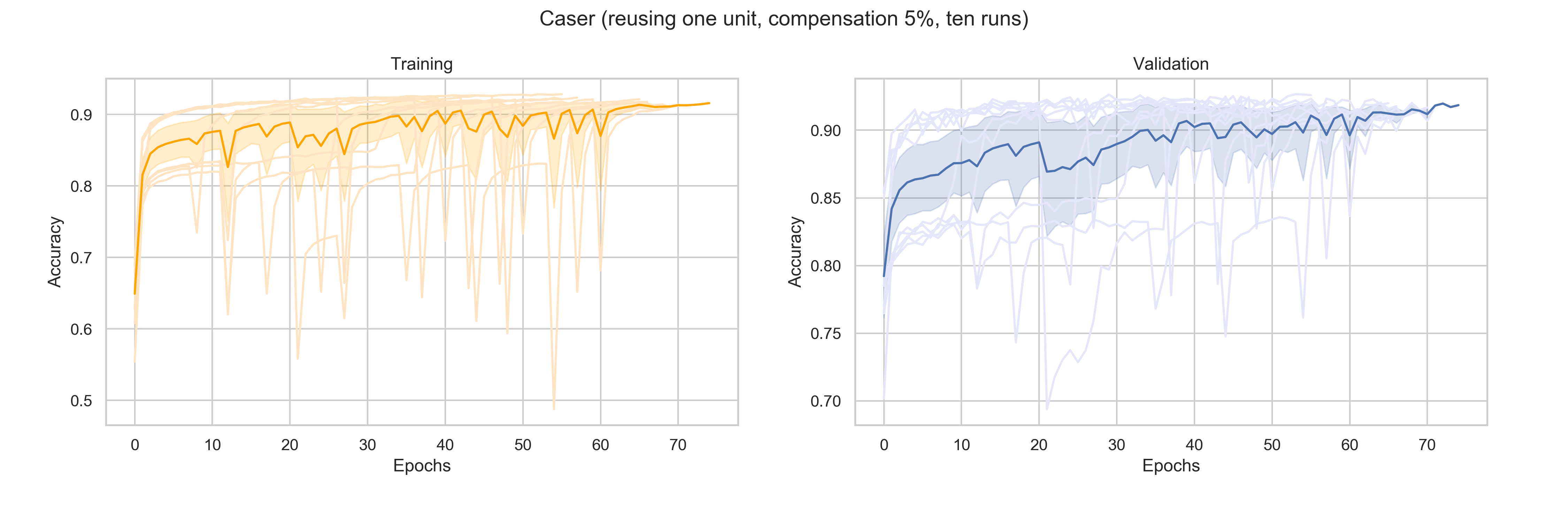}
  \caption{Caser, reusing the previous output weight vector if all units in the candidate pool decrease the networks accuracy by more than 5\%.}
  \label{fig:caser-re1-comp}
\end{figure}

Figure \ref{fig:caser-re1-comp} reuses the previous output weight vector if the new unit decreases the validation accuracy by more than 5\%. The overall performance of the network is improved, however, the figure shows some drastic performance drops during training.

\begin{figure}[h]
  \center
  \includegraphics[width=\textwidth]{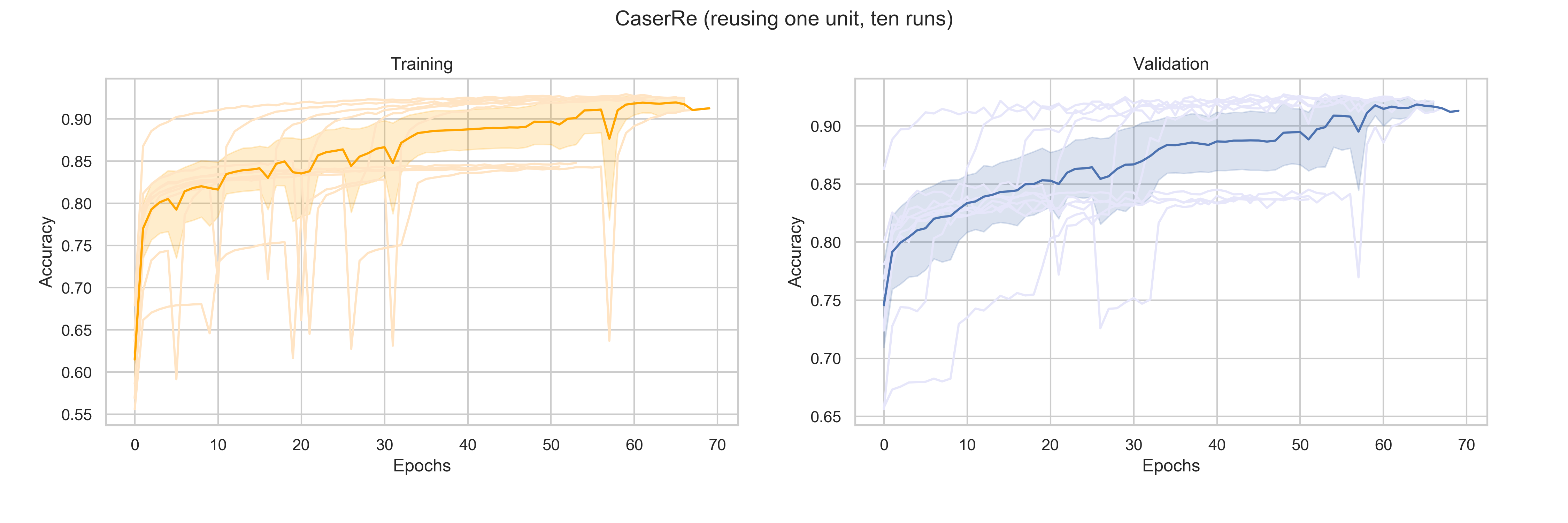}
  \caption{Using a candidate pool of seven new units and one unit reusing the previous output weights. Results averaged over ten runs, with the shaded area representing a 95\% confidence interval. Lighter colored lines show the single runs.}
  \label{fig:caser-re3}
\end{figure}

Another approach is to modify the candidate pool. Instead of training eight candidate units, we can train seven new candidate units and one candidate unit that reuses the previous output weights. In this way, we will only change the output weights if it leads to an increase in test accuracy. Obviously, the newly trained units will only be trained for one epoch while the unit reusing output weights has been trained to convergence. To make up for this difference, we could set a compensation factor. In the experiments plotted in Figure \ref{fig:caser-re3}, I did not use such a compensation factor for the sake of automaticity (the fewer tunable parameters, the better).

This shows good results, with the network reaching an accuracy of over 90\% in 7 out of 10 training runs, with the remaining 3 runs achieving an accuracy of over 83\%. 

\begin{figure}[h]
  \center
  \includegraphics[width=\textwidth]{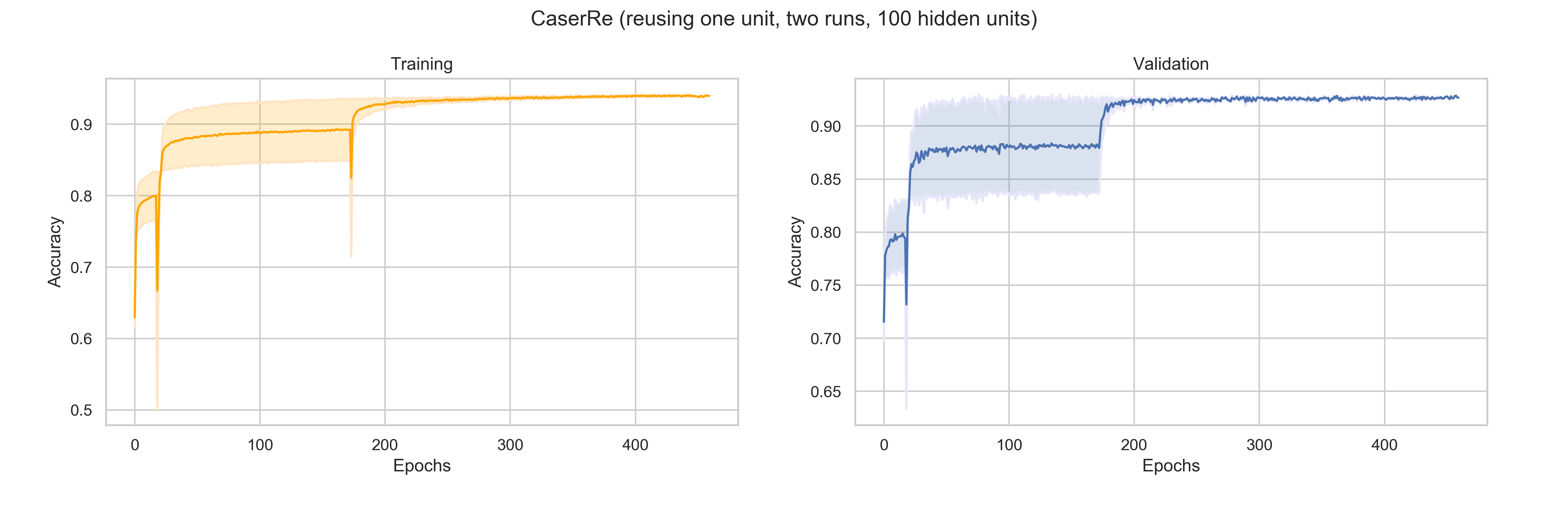}
  \caption{Using a candidate pool of three new units and one unit reusing the previous output weights. Adding a total of 100 cascading hidden units. Results averaged over two runs, with the shaded area representing a 95\% confidence interval. Lighter colored lines show the single runs.}
  \label{fig:caser-max100}
\end{figure}

So far, it seems like all experiments on Cascor, Caser, and CaserRe have been underfitting on the MNIST learning task, as they have been using only ten hidden units in total - as compared to standard MLPs that have hundreds of hidden units. I trained the algorithm whose results are shown in Figure \ref{fig:caser-re3} for 100 cascading hidden units for two training runs, using a candidate pool size of four. The results are shown in Figure \ref{fig:caser-max100}; both networks reach a validation accuracy of 92.7\%. 

A comparable MLP with one hidden layer of 100 neurons reaches a validation accuracy of around 94.0\% (trained with RMSProp on crossentropy loss, using tanh in hidden units and softmax in the output layer).This shows that CaserRe is close to the performance of comparable layered networks. However, in order to be competitive on the MNIST learning task, a testing accuracy of over 95\% should be achieved. The complexity of the CaserRe network needs to be increased in an attempt to learn the MNIST task to a higher accuracy. The insertion of hidden units is computationally expensive due to the training of the candidate pool and modifications to the computational graph of the neural network. Complexity may be added into the network more efficiently by increasing the complexity of each hidden unit, e.g. by replacing a hidden \emph{unit} by a hidden cascading \emph{layer}. To the best of my knowledge, this has not been done before.

\begin{figure}[h]
  \center
  \includegraphics[width=\textwidth]{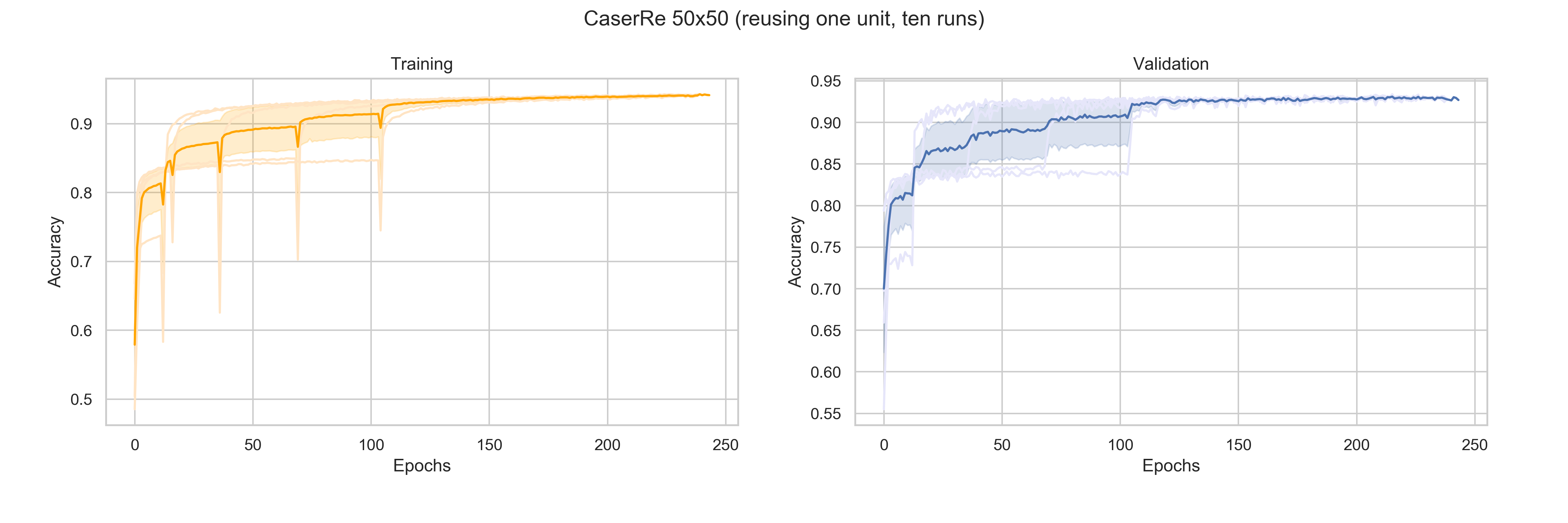}
  \caption{Using a candidate pool of three new units and one unit reusing the previous output weights. Adding a total of 50 cascading hidden layers of 50 units each. Results averaged over five runs, with the shaded area representing a 95\% confidence interval. Lighter colored lines show the single runs.}
  \label{fig:caser-50x50}
\end{figure}

I ran another experiment, using candidate layers rather than single candidate units. Each candidate layer contains 50 neurons and a total of 50 of these cascading layers were inserted into the network. I used a candidate pool of size four. The result is shown in Figure \ref{fig:caser-50x50}, the network reaches a validation accuracy of 92.85\% (averaged over five runs with a standard deviation of 0.20\%). This is slightly better than the Caser architecture with 100 cascading hidden units and worse than layered networks of similar architecture.

\begin{figure}[h]
  \center
  \includegraphics[width=\textwidth]{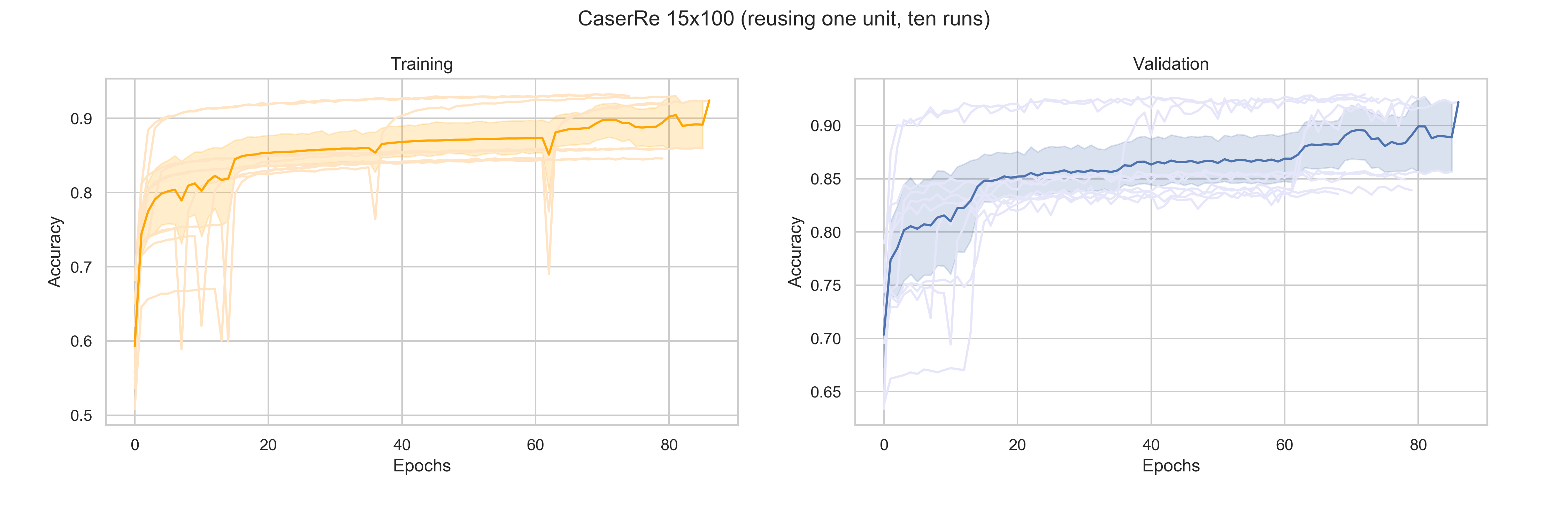}
  \caption{Using a candidate pool of three new units and one unit reusing the previous output weights. Adding a total of 15 cascading hidden layers of 100 units each. Results averaged over five runs, with the shaded area representing a 95\% confidence interval. Lighter colored lines show the single runs.}
  \label{fig:caser-15x100}
\end{figure}

In another experiment, I used layers of size 100, adding a total of 15 of these cascading layers into the network - again using a candidate pool size of four. The results for this architecture are shown in Figure \ref{fig:caser-15x100}. The network reaches a validation accuracy of 88.58\% (averaged over ten runs and a standard deviation of 4.11\%) with a maximum accuracy of 92.93\% and a minimum of 83.57\%. Again, this is worse than comparable layered architectures.

Even though the resulting networks are very large, they do not overfit on the MNIST dataset and the final performance does not significantly change when adding more complexity into the network (by adding cascading layers rather than single units to increase the width or by increasing the depth of the cascading units/layers). A more detailed investigation into the connection weight values from the hidden activation vector compared to the input activation vector may bring some insights. In order to prioritize the cascaded hidden units/layers over the input vector, one may drop out or reduce some of the input-to-output connection weights (through dropout or weight decay) in order to incentivize the network to make more use of the new hidden activation vector.

\paragraph{Cascor Summary}\mbox{}\\

After some additional work based on Cascor and Caser, I was able to find a well-performing learning algorithm, which I called CaserRe. Though the final algorithm is able to find good local minima with an average accuracy of over 90\%, adding more units and layers into the network does not increase performance to anything above 93\% testing accuracy.

One reason for this may be that the input to each subsequent hidden unit is still very noisy. Traditional layered neural networks map the input to a different dimension through the first hidden layer. Subsequent hidden layers work only on the output of previous layers. Hidden layers could be seen as making the data more well-behaved, as suggested by \cite{forwardthinking}. This may be why the forward thinking algorithm seems to work much better than my current implementation of different Cascor versions which are facing problems with the aforementioned volatility.

Another way to look at is that the error surface (with respect to the weights) is very high dimensional, as the weight vector is very high dimensional. With each added unit, the network tries to find a new local minimum, with one weight being fixed (i.e. one degree of freedom on the error surface frozen) and the rest still to be varied. Since the input dimension is much higher than the dimension of all hidden units combined (in my experiments, no more than one hundred hidden units/layers have been inserted into the network while the input layer has over 700 units), the error minimization problem is dominated by the connections weights from the input to the output. In order for this issue to disappear, one would have to train a \emph{very} deep cascading network in order for the hidden weights to be more important in relation to the input-to-output-connection weights. This would explain why Cascor performs well on datasets with lower dimensionality, such as the problems treated in the original paper, because there the input-to-output-connection weights are much fewer and thus less relevant in comparison to the hidden weights. 

In terms of performance, training these cascading networks can be very efficient using modern deep learning frameworks, with each epoch taking no more than a few seconds. However, the cascading structure requires making changes to the computational graph, which sum up to be a large overhead. The deeper networks (50 cascading layers of 50 units each, 100 cascading layers of single units, and 100 15 cascading layers of 100 units each) took over 30 minutes to train, with the vast majority of the time spent on the training of candidate unit/layers. This can be done much more efficiently, since the candidate training allows for perfect parallelization. Hence the candidate unit training can be done in parallel and yield a time decrease of up to 8x.

Since most modern neural networks deal with very high-dimensional data, more work on Cascor is required in order to make it competitive in the world of modern neural networks. A comprehensive study on different cascading architectures can give more conclusive evidence for whether or not these cascading architectures can perform as well, or better, compared to similar layered architectures.

\subsubsection{Forward Thinking}

The forward thinking algorithm trains a fully-connected neural network by building up the network one hidden layer at a time \citep{forwardthinking}. The originally proposed algorithm does not automate the layer construction. One needs to specify how many layers can be added, as well as the width of the layer and the activation function used. The networks in my experiments will be trained on cross entropy loss using RMS Prop. Hidden units use the tanh or relu activation function, output units use softmax.

Parameters that needed to be decided on include:
\begin{itemize}
  \item Hidden layers: how many layers, how many units in each layer, activation functions.
  \item Layer construction time: when to add new layers.
\end{itemize}

For this experiment, a new layer will be added when the training of the current layer has not improved the accuracy on the validation data for two epochs in a row (and training will be stopped after the validation accuracy hasn't improved for three epochs in a row when training the last layer). I am running the forward thinking algorithm on three different architectures: two layers of 512 tanh units each, three layers of 850 tanh units each, and five layers of 750 tanh units each - taking the best-performing neural network setups from the random search results using two, three and five hidden layers. 

\begin{figure}[h]
  \center
  \includegraphics[width=\textwidth]{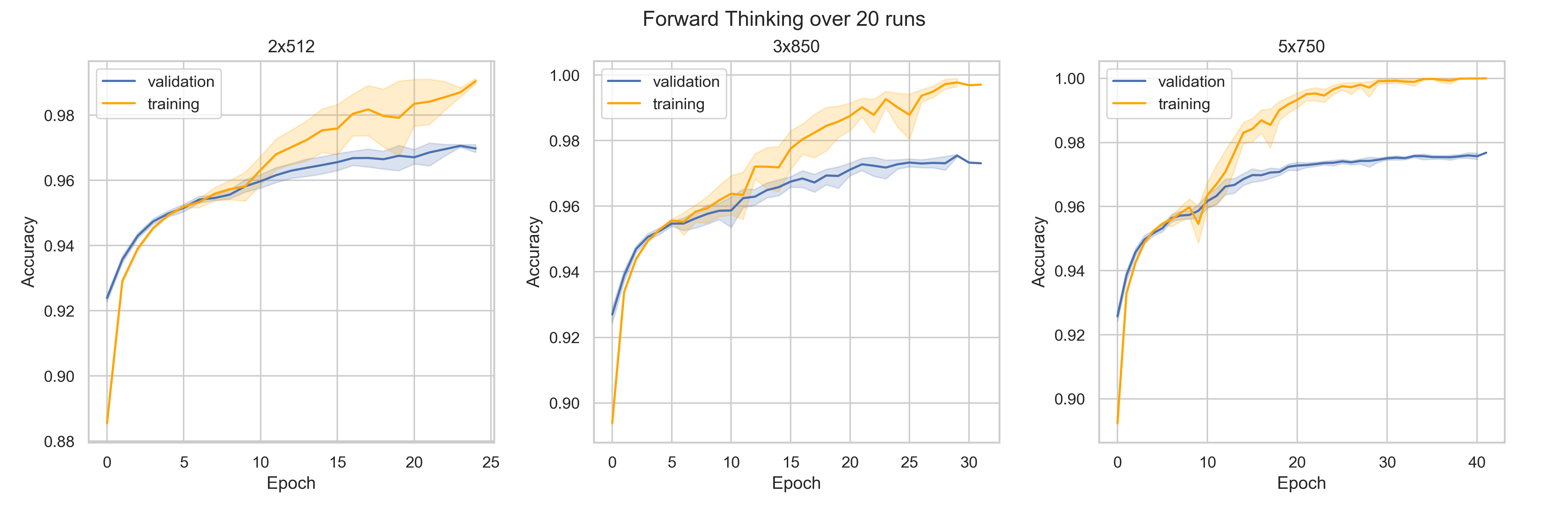}
  \caption{Training and validation accuracy per epoch in forward thinking. Results are averaged over 20 runs, the shaded areas show the 95\% confidence interval.}
  \label{fig:ft-acc-pat2-all}
\end{figure}

Figure \ref{fig:ft-acc-pat2-all} shows the performance of these networks. It is interesting to see that the testing accuracy seems to reach its maximum around half-way through each layer-wise training (or even slightly before) while the training accuracy continuously increases. Moreover, while the training accuracy decreases significantly when a new layer is inserted, the testing accuracy does not suffer from this decrease. Near the training's end, the training accuracy keeps increasing significantly more than the validation accuracy. This looks strange - it doesn't seem to be overfitting, as the validation accuracy keeps improving as well. This is very similar to the findings reported by \cite{forwardthinking}.

\begin{figure}[h]
  \center
  \includegraphics[width=\textwidth]{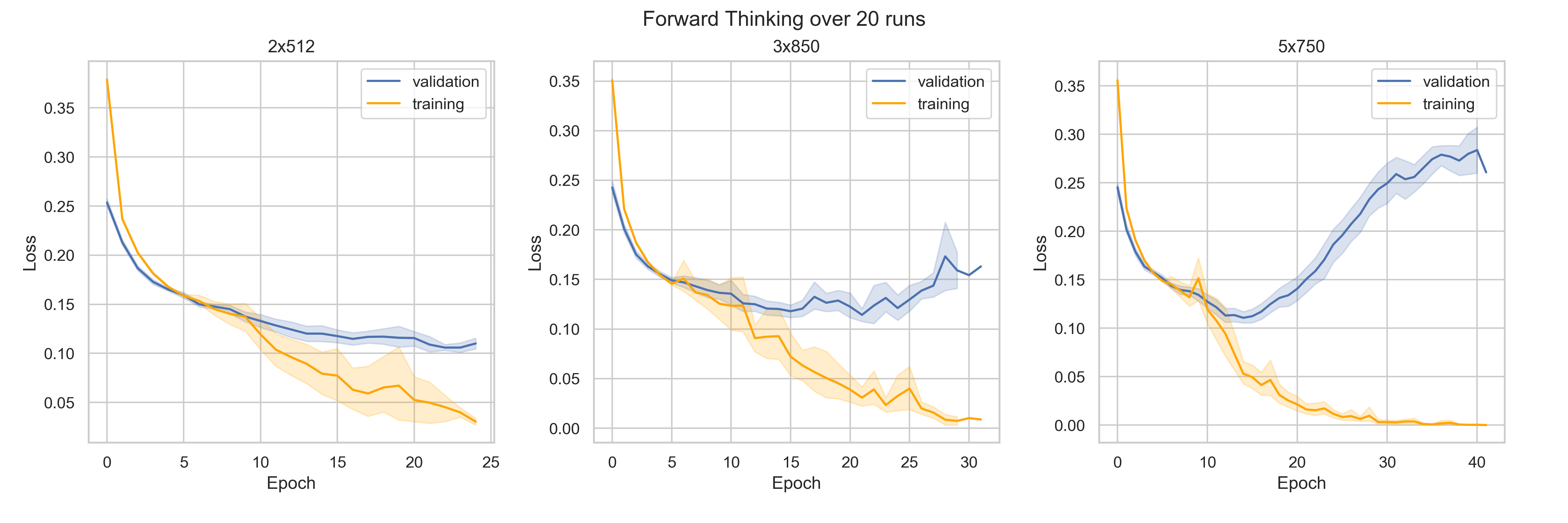}
  \caption{Training and validation loss per epoch in forward thinking. Results are averaged over 20 runs, the shaded areas show the 95\% confidence interval.}
  \label{fig:ft-loss-pat2-all}
\end{figure}

However, looking at the loss, shown in Figure \ref{fig:ft-loss-pat2-all}, demonstrates that the network is indeed starting to overfit, but the accuracy doesn't suffer from the overfitting. This effect is more significant in deeper networks. 

\cite{forwardthinking} do not report the loss of their network, hence a direct comparison is not possible. The accuracy is computed through an argmax operation on the output vector (see Appendix \ref{ss:loss-functions}). As long as the maximum value in the output vector belongs to the same class, the accuracy does not change. However, if the output vector becomes less certain about the class - meaning that the difference between the maximum argument and other arguments decreases - the loss will increase, penalizing this increased uncertainty. Hence, the forward thinking algorithm is indeed starting to overfit on the training data, with the overfitting being more significant in deeper networks.

Early stopping on the accuracy doesn't seem to avoid overfitting as well as early stopping on the loss would. Hence, the following experiments will be applying early stopping to the validation loss, rather than the validation accuracy.

The final performance of these networks is shown in Table \ref{tbl:forward-thinking} and for a direct comparison between forward thinking and "standard" training, the same statistics are shown in Table \ref{tbl:forward-thinking-comparison} for a network trained using backpropagation. 

\begin{table}[h]
\begin{center}
\begin{tabular}{| c | c | c | c | c |}
  \hline
  Layers & Epochs & Train Accuracy & Validation Accuracy & Time\\
  \hline
  2 x 512 & 18 $\pm$ 4 & 98.75\% $\pm$ 0.48\% & 96.85\% $\pm$ 0.28\% & 28.9s $\pm$ 6.5s \\
  3 x 850 & 21 $\pm$ 4 & 99.30\% $\pm$ 0.27\% & 97.27\% $\pm$ 0.23\% & 35.7s $\pm$ 5.8s \\
  5 x 750 & 29 $\pm$ 4 & 99.91\% $\pm$ 0.08\% & 97.54\% $\pm$ 0.12\% & 47.4s $\pm$ 5.7s \\
  \hline
\end{tabular}
\end{center}
\caption{Network performances when trained with forward thinking. Results show the averages and standard deviations over 20 training runs.}
\label{tbl:forward-thinking}
\end{table}

\begin{table}[h]
\begin{center}
\begin{tabular}{| c | c | c | c | c |}
  \hline
  Layers & Epochs & Train Accuracy & Validation Accuracy & Time\\
  \hline
  2x512 & 23 $\pm$ 6 & 98.45\% $\pm$ 0.24\% & 97.27\% $\pm$ 0.15\% & 36.5s $\pm$ 8.8s \\
3x850 & 18 $\pm$ 5 & 98.09\% $\pm$ 0.19\% & 96.92\% $\pm$ 0.26\% & 36.2s $\pm$ 9.2s \\
5x750 & 20 $\pm$ 6 & 97.19\% $\pm$ 0.23\% & 96.10\% $\pm$ 0.32\% & 48.4s $\pm$ 14.8s \\
  \hline
\end{tabular}
\end{center}
\caption{Network performances when trained using backpropagation (for a direct comparison between backpropagation and forward thinking. Results show the averages and standard deviations over 20 training runs.}
\label{tbl:forward-thinking-comparison}
\end{table}

The results show that the two layer network performs 0.4\% better (on average) when trained using backpropagation. The three and five layer networks show a 0.3\% and 1.5\% increase in validation accuracy (on average) when trained with forward thinking. This is in agreement with forward thinking being more efficient in training deep neural networks, as there is no need to propagate the error signal through many layers. More experiments on other learning tasks are needed in order to solidify this hypothesis.

\cite{forwardthinking} reported a 30\% decrease in training time on a four-layer neural network. Though forward thinking was, on average, faster for all three network architectures, I cannot report the same magnitude of speedup. This may be due to the fact that the training happens on the GPU but the computational graph is modified after each layer-wise training which entails that data has to be moved to and from the CPU. This leads to a larger overhead in computation, as previously mentioned for the cascading networks in Section \ref{ss:exp-cascor}. In order to test this hypothesis, I ran the same experiment on my personal computer's CPU (running the training once for backpropagation and once for forward thinking due to time constraints). This indeed shows a much larger improvement in training time for forward thinking compared to backpropagation - 46\% for the 5 x 750 network, 25\% for the 2 x 512 network and 53\% for the 3 x 850 network. The test accuracy is similar to the ones reported previously. The result is shown in Table \ref{tbl:ft-backprop-appendix} in Appendix \ref{ss:appendix-results}.

\subsubsection{Automated Forward Thinking}

In order to automate forward thinking more, one might want to automate the choice of layers that will be added into the network. Inspired by the original Cascor algorithm \citep{cascor}, I use a pool of candidate layers - training each one for a few epochs and choosing the best layer from the candidate pool to insert into the network. To the best of my knowledge, this has not been done before. 

\begin{figure}[h]
  \center
  \includegraphics[width=0.5\textwidth]{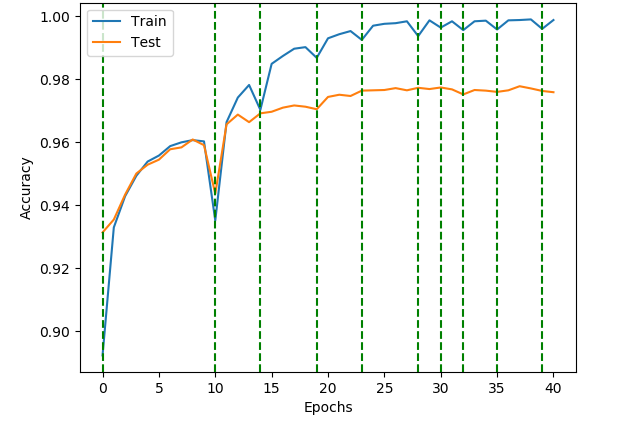}
  \caption{The automated forward thinking algorithm, trained for ten layers. Resulting network has the layers: [950, 700, 700, 500, 50, 200, 500, 850, 550, 350].}
  \label{fig:auto-ft-10epochs}
\end{figure}

In my experiments, I used a candidate pool of eight layers, each layer being trained for two epochs. The width of each candidate layer is chosen at random within the interval $[50, 1000]$, restricted to multiples of 50. The best performing of these eight candidate layers will be inserted into the network and fully trained (until the validation accuracy stops improving). This already works reasonably well, as shown in Figure \ref{fig:auto-ft-10epochs}. However, not all layers are needed for the final model to perform as well as it does. The first two layers offer significant increases in accuracy, but this increase in model performance flattens quickly. A stopping criterion which detects this performance flattening could yield smaller networks with similar performance. 

\begin{figure}[h]
  \center
  \begin{subfigure}[b]{0.48\textwidth}
    \includegraphics[width=\textwidth]{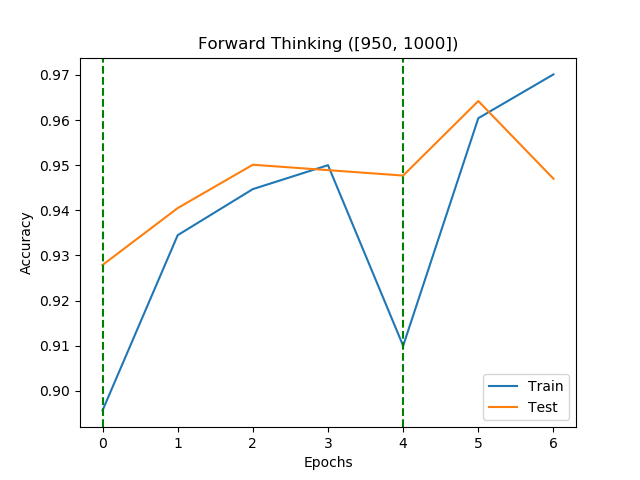}
   \caption{Training stopped too early.}
    \label{fig:auto-ft-stopping1-too-soon}
  \end{subfigure}
  \begin{subfigure}[b]{0.48\textwidth}
    \includegraphics[width=\textwidth]{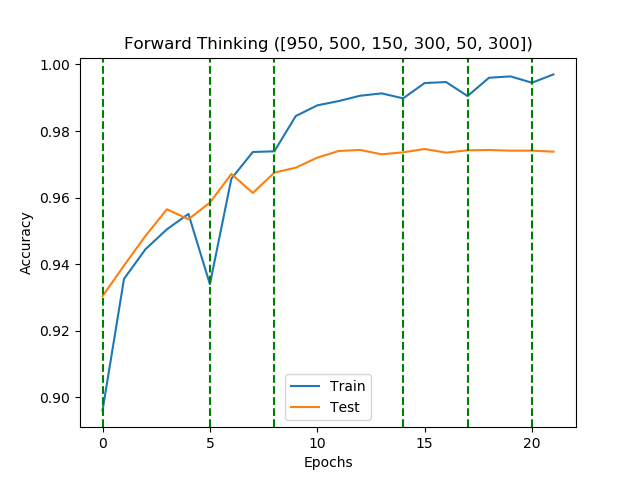}
   \caption{Training stopped too late.}
    \label{fig:auto-ft-stopping1-too-late}
  \end{subfigure}
  \caption{Automated forward thinking with early stopping when the validation accuracy does not increase after adding a layer. The network on the left has two layers: [950, 1000], whereas the network on the right has six layers: [950, 500, 150, 300, 50, 300].}
  \label{fig:auto-ft-stopping1}
\end{figure}

The stopping criterion is an opportunity to automate the algorithm further. Early stopping seems to be a reasonable choice. I ran some experiments using early stopping, which ends the training when the layer's final validation accuracy hasn't improved over the previous layer's final validation accuracy. Figure \ref{fig:auto-ft-stopping1} shows that this approach is not ideal. In Figure \ref{fig:auto-ft-stopping1-too-soon}, one can argue that training was stopped too early, the network could have improved further, whereas in Figure \ref{fig:auto-ft-stopping1-too-late}, training was stopped too late, adding more layers than necessary as one can see from the flattened training accuracy after the fourth layer was inserted. It might help to train each layer for a longer time, in order to have a more reliable value for the final layer's validation accuracy. 

Early stopping is commonly used to stop training neural networks of fixed architectures and to avoid overfitting. Normally, the penalty of training a neural network for one (or a few) epochs is not very high. However, the penalty of adding one (or a few) layers more into a neural network is very large - the complexity of the resulting model increases substantially. A stricter version of early stopping is needed. 

\begin{table}[h]
\begin{center}
\begin{tabular}{| c | c | c | c | c | c |}
	\hline
	Layers & Test Acc & Train Acc & Total & Train & Layers \\
	\hline
	4 & 97.86\% & 99.95\% & 185.16s & 99.33s & [900, 600, 600, 300] \\
	4 & 97.68\% & 100.00\% & 180.60s & 99.77s & [700, 700, 400, 300] \\
	4 & 97.68\% & 100.00\% & 184.44s & 96.82s & [900, 900, 300, 300] \\
	4 & 97.64\% & 99.99\% & 184.97s & 103.40s & [900, 500, 400, 200] \\
	4 & 97.51\% & 99.99\% & 146.33s & 66.36s & [800, 600, 100, 100] \\
	4 & 97.47\% & 99.53\% & 163.28s & 84.37s & [1000, 200, 100, 100] \\
	3 & 97.46\% & 100.00\% & 148.40s & 91.36s & [1000, 200, 100] \\
	3 & 97.44\% & 99.90\% & 140.55s & 83.30s & [900, 100, 100] \\
	3 & 97.30\% & 100.00\% & 144.90s & 86.53s & [600, 500, 300] \\
	3 & 97.16\% & 99.62\% & 112.55s & 55.35s & [800, 100, 100] \\
	\hline
\end{tabular}
\end{center}
\caption{Ten smallest architectures found by running the automated forward thinking algorithm 20 times. Train gives the actual training duration, while Total gives the total training time, including the candidate unit training.}
\label{tbl:aft-best}
\end{table} 

\begin{figure}[h]
  \center
  \includegraphics[width=\textwidth]{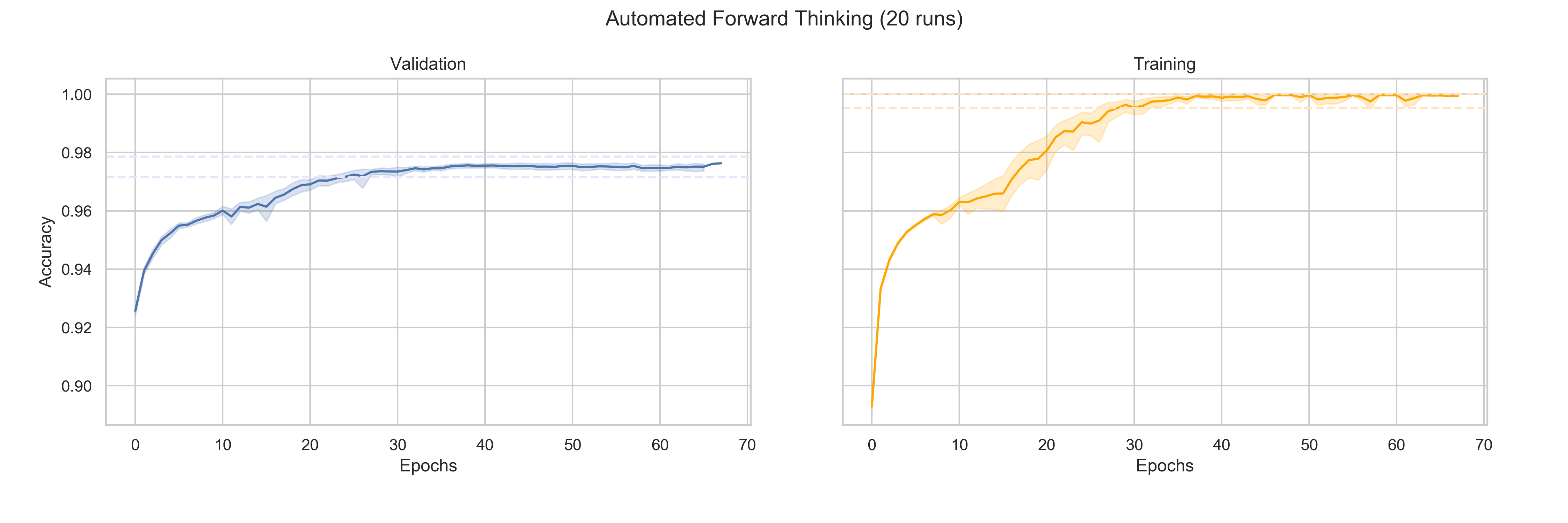}
  \caption{The automated forward thinking algorithm run 20 times. Shaded area shows the 95\% confidence interval.}
  \label{fig:auto-ft-20runs}
\end{figure}

Considering that the training using forward thinking is quite fast, it is computationally feasible to insert more layers into the network than needed, storing the network performance for all number of layers. Based on this, one may assess with how many layers the training reaches an optimal tradeoff of performance against model complexity. Finally, unnecessary layers can be removed from the network and the output weight vector can be retrained. I implemented and ran the algorithm 20 times, yielding 20 unique architectures. I furthermore restrict the algorithm to only use layers of subsequently decreasing widths as that is how most neural network architectures are designed. This decision is subject to more discussion, though I will ommit this discussion in my thesis. Table \ref{tbl:aft-best} shows all architectures using fewer than five layers. Figure \ref{fig:auto-ft-20runs} shows the training graph for this.

Across 20 runs of the algorithm, the average test accuracy is 97.54\% (with a standard deviation of only 0.17\%) - which is better than any other algorithm I have investigated in this thesis. Half of the architectures use below five layers, the other half uses five or more layers. The best performing network architecture is [900, 600, 600, 300] with a testing accuracy of 97.86\%.

The increased performance over layered neural networks likely stems from the difficulty of training deep networks with backpropagation. Training the network using forward thinking may enable the algorithm to take deeper, more complex architectures into consideration and train them more efficiently than backpropagation could. 

\subsubsection{Conclusion}

In this section of constructive dynamic learning algorithms, I compared cascading networks and forward thinking networks, each being a category of several learning algorithms. The most promising algorithms are CaserRe, forward thinking and automated forward thinking. However, as forward thinking does not design its own architecture - it is an algorithm to train a neural network - I will not be considering it as an automated architecture design algorithm. 

In terms of automation, both automated forward thinking and CaserRe show a similar level of automaticity. Both algortihms search for a suitable architecture automatically, in a randomized greedy way through the use of a candidate pool. Automated forward thinking needs an upper and lower bound for the hidden layers' widths. In CaserRe, one also needs to specify whether hidden units or hidden layers should be inserted in a cascading way (and how large these hidden layers may be). 

Automated forward thinking outperforms CaserRe in the MNIST learning task by 5\% on the testing accuracy (CaserRe with 50 cascading hidden layers of 50 units each). 

The automated forward thinking and CaserRe algorithms have very similar computational requirements (given the same candidate pool sizes). However, CaserRe needs to add more cascading units (or layers) into the network than automated forward thinking needs to add layers, hence CaserRe could be said to be slower than automated forward thinking. However, as there is a significant performance difference between the two algorithms, no exact comparison in terms of computational requirements can be made.

The resulting model complexity of automated forward thinking networks and CaserRe networks is difficult to assess, as there is a performance difference between the two and because I have no basis for comparing layered networks with cascading networks - other than the empirical evidence that cascading networks do not seem to be able to learn the MNIST learning task as well as automated forward thinking.

In summary, CaserRe is in need of further investigation in order to get its performance levels to competitive standards, or in order to explain why this cascading structure may not be suitable for a learning task such as MNIST. Automated forward thinking seems to be a very well-performing constructive learning algorithm, outperforming all neural networks trained using standard backpropagation that I covered in this thesis. Further empirical evidence is needed to confirm the experimental results from my work in this thesis.

\subsection{Conclusion}

The empirical investigation laid out in this thesis give a preliminary overview of some techniques for the automated architecture design of deep feedforward neural networks. Good results have been reported and preliminary hypotheses about the suitability of different algorithms have been made.

The experimental findings show that different neural architecture search algorithms are able to find suitable network architectures that perform well on the learning task. The neural architecture search investigation hints at possible use cases to search for well-performing architectures. Manual search is best used when a lot of knowledge about good architectures is available, either through experience or through available results in the literature. Random search can be used to evenly explore the search space, if the goal is to explore the entire search space without any bias introduced through prior knowledge. Evolutionary search strikes a compromise between the unbiasedness of random search and the manual search algorithm driven primarily by prior (human) knowledge.

Furthermore, as constructive dynamic learning algorithms, this thesis includes a preliminary investigation of two families of such algorithms: the recently proposed forward thinking algorithm and the cascade-correlation learning architecture that was proposed over twenty years ago. Both algorithms have been implemented on the digit classification learning task. I extended both algorithms to improve their performance and level of automaticity. Results have been reported on the learning task and the algorithms' merits have been discussed. The investigated cascading architectures were not able to perform as well as standard layered networks - more work is needed to assess, and possibly enhance, their viability on modern learning tasks. The forward thinking algorithm outperformed all layered neural networks investigated in this thesis and shows promise for future work, despite more work being needed on regularizing this architecture in order to combat overfitting and improve generalization. 

Automated forward thinking extends the greedy-wise training proposed by forward thinking into a fully automated architecture design algorithm for neural networks. The algorithm builds a network deeper than the standard MLP architectures found with the search algorithms described above and yields better performance on the test data than any MLP investigated in this thesis. As such, automated forward thinking shows a promising technique that may further be investigated in more comprehensive studies.

To summarize, this thesis has given a preliminary overview of exisiting algorithms for the automation of architecture design and reported some results on a selected learning task of digit classification. The results of this thesis may be used as a starting point for further work on fully, and partially, automated architecture design algorithms for deep neural networks. If the trend of creating more and more complex deep learning models continues, these automated architecture design algorithms may be the main tools to design neural networks for new learning tasks in the future.
  \section{Future Work}

As stated previously, this thesis merely gives a preliminary overview of automated architecture design algorithms for deep feedforward neural networks and empirical results to guide the direction of future research. Possible future research directions in the field of automated architecture design are outlined in this section.

The first large restriction of this research project is the limitation to feedforward neural networks. Future research may investigate techniques for the automated architecture design of other types of neural networks, most notably convolutional neural networks and recurrent neural networks. The original forward thinking algorithm has also been applied to convolutional neural networks \citep{forwardthinking} and a recurrent version of cascade-correlation neural networks was proposed by \cite{recascor}. 

Neural architecture search has already been applied to a large variety of different neural networks. For example, \cite{real2018regularized} evolved a neural network architecture that ultimately outperformed manually crafted architectures for the first time on the ImageNet learning task. They are using the NASNet search space for the evolution of their architecture that was designed by \cite{Zoph_2018_CVPR}. \cite{real2018regularized} further also compared their evolutionary search with different neural architecture search algorithms, specifically with random search and reinforcement learning applied to neural network architectures. Future work in the field may run more comparative studies on neural architecture search algorithms, establishing some empirical evidence for the circumstances under which each neural architecture search algorithm performs well. 
Moreover, an in-depth analysis of different neural architecture search based on the properties of the search space may be able to establish some formal proofs or evidence of certain search algorithms being more advantageous than others, for different kinds of learning tasks. Such a general analysis is inherently difficult and may only be possible after comprehensive empirical evidence is available on a large set of diverse learning tasks. The survey provided by \cite{elsken2018neural} on neural architecture search algorithms may be a starting point for such in-depth, largely task-independent research.

Neural networks that change their network architecture based on the learning task, i.e. learning both the architecture and the connection weights simultaneously have not been worked on in the same magnitude as the field of neural architecture search, to the best of my knowledge. This may be due to the lack of a unifying term of such algorithms. \cite{waugh1994dynamic} uses the term \emph{dynamic learning} for such models, \cite{cortes2017adanet} uses the term \emph{adaptive structural learning}, and \cite{evolvinganns} uses the term \emph{evolving ANNs} for neural networks whose architecture and parameters are learned simultaneously using evolutionary search algorithms. One term that may contain all these terms is \emph{automated machine learning}, or AutoML. However, I was not able to find such a term specifically for neural networks, which could be seen as a subset of AutoML. Moreover, the most recent survey of such models that I was able to find at the beginning of this research project was over 20 years old, by \cite{waugh1994dynamic}. In April 2019, \cite{zoller2019survey} submitted a survey on automated machine learning to the Journal of Machine Learning Research (pending review). The survey gives  a good overview of recent work in the field of automated machine learning, but I found it to not be comprehensive with respect to automated architecture design for neural networks, as its focus lies more in the automation of the entire machine learning pipeline. As automated machine learning can be seen as a superset of automated architecture design for neural networks, the survey is still highly relevant but not comprehensive. Future work in the field of automated neural network architecture design should include a survey that gives an overview of the most relevant techniques - techniques that learn both the architecture and the paramters of the networks simultaneously.

The future work in the field of automated architecture design for neural network with I am proposing in this thesis can be summarized as (1) compiling a survey of the most relevant techniques for automated architecture design, (2) gathering empirical evidence for the performance and comparison of different algorithms on diverse learning tasks, and (3) establishing formal proofs or concrete evidence for task-independent performance of different algorithms.

  \appendix
\addtocontents{toc}{\protect\setcounter{tocdepth}{0}}

\section{Appendix}

\subsection{Loss functions}
\label{ss:loss-functions}

\subsubsection{Crossentropy Loss}

The crossentropy loss is a loss function for multi-class classification problems. The categorical cross-entropy loss refers to the use of the softmax activation function on the output and then the cross-entropy loss.

Let $N$ be the number of patterns in the dataset, $C$ the number of classes, and $p_{model}(y_i \in C_c)$ is the probability given by the model that pattern $i$ belongs to the class $c$.

$$
- \frac{1}{N} \sum_{i=1}^N \sum_{c=1}^C 1_{y_i \in C_c} \log p_{model}(y_i \in C_c)
$$

where 

$$
1_{y_i \in C_c} = \begin{cases}
  1 & y_i \in C_c\\
  0 & y_i \notin C_c
\end{cases}
$$

\subsubsection{Error Correlation Maximization}
\label{ss:error-correlation}

The error correlation maximization was proposed to train cascade-correlation neural network by \cite{cascor}. The objective of the algorithm is to maximize the error correlation $S$, which is given by:

$$
S = \sum_{o \in O} \left| \sum_{p \in P} (V_p - \bar V) (E_{p,o} - \bar E_o) \right|
$$

where $O$ is the set of output units and $P$ is the training dataset. $V_p$ is the hidden unit's value (its activation) when the training pattern $p$ was passed through the network. $\bar V$ is the hidden unit's value averaged over all training patterns. $E_{p,o}$ is the error at the output unit $o$ on the training pattern $p$ and $\bar E_o$ is the error at output unit $o$ averaged over all training patterns. 

\subsubsection{Accuracy Computation}
\label{ss:accuracy}

In the experiments contained in this thesis, the primary performance metric is the accuracy of the neural network's predictions on a classification task. Let $C$ be the set of $|C|=c$ classes. Let the output of the neural network be given by $y$ where $y \in \mathbb{R}^c$. After passing the output of the neural network through the softmax function $\sigma$, we obtain $z = \sigma(y)$ where $z \in \mathbb{R}^c$. The accuracy $\tau$ can be computed as follows:

$$
\tau = \text{argmax}_{i} \ z
$$

where $i \in \{1, \ldots, c\}$.

\subsection{Activation functions}
\label{ss:activation}

In my thesis, I am using three different activation functions, namely relu (Rectified Linear Unit), tanh (hyperbolic tangent), and softmax.

The relu function is a function $\text{relu}: \mathbb{R} \rightarrow \mathbb{R}$:

$$
\text{relu}(x) = \begin{cases}
  0 & x < 0 \\
  x & x
\end{cases}
$$

The tanh function is a function $\text{tanh}: \mathbb{R} \rightarrow \mathbb{R}$:

$$
\text{tanh}(x) = \frac{e^x - e^{-x}}{e^x + e^{-x}} \in [-1, 1]
$$

Both the relu and the tanh function can be applied to vectors of real numbers by applying the function to each of its elements individually.

The softmax function $\sigma$ is defined on a vector of $K$ real numbers and normalizes that vector into a probability vector, $\sigma: \mathbb{R}^K \rightarrow \mathbb{R}^K$:

$$
\sigma(z)_i = \frac{e^{z_i}}{\sum_{j=1}^K e^{z_j}} \in [0,1]
$$

where $z \in \mathbb{R}^K$ and $1 \leq i \leq K$. 

\subsection{Neural Network Optimization Algorithms}
\label{s:nn-optimization-algorithms}

This section closely follows and paraphrases the paper by \cite{ruder2016overview} which gives a good overview of different gradient descent optimization algorithms commonly used for training neural networks.

\subsubsection{Stochastic Gradient Descent}

There are different variations of the standard gradient descent algorithm that vary in the amount of data that they take in before updating the parameters.

Let $N$ be the number of patterns in the training data, $\eta$ be the learning rate, $\theta$ be the parameter vector (the vector of all connection weights in a neural network), $\mathcal{L}_i(\theta)$ be the loss for pattern $i$ (given parameter vector $\theta$), then the standard ("batch") gradient descent algorithm updates the weight vector in the following way:

$$
\theta_{t+1} = \theta_t - \eta \frac{1}{N} \sum_{i=1}^N \nabla \mathcal{L}_i(\theta_t)
$$

where $t$ indicates the step of the gradient descent optimization.

This computation can be slow and for large datasets even intractable if they do not fit into memory. We can break down the update rule and update the parameter vector with every single pattern that we train on. This is called stochastic gradient descent, which is applied for every pattern $i \in \{1, \ldots, N \}$:

$$
\theta_{t+1} = \theta_t - \eta \nabla \mathcal{L}_i(\theta_t)
$$

However, this is not very efficient either, because we update the parameter vector for every single pattern in the dataset. In order to strike a compromise between batch gradient descent and stochastic gradient descent, one may use so-called "mini-batches", i.e. subsets of the total training data of size $m$, after each of which the parameters are updated as follows:

$$
\theta_{t+1} = \theta_t - \eta \frac{1}{m} \sum_{i=1}^m \nabla \mathcal{L}_i(\theta_t)
$$

This is called mini-batch gradient descent and it is the algorithm that I am referring to as SGD (stochastic gradient descent) because this is what the algorithm is called in Keras, the deep learning framework that I am using for my code implementations. For all experiments found in this thesis, I used a mini-batch size of 128.

\subsubsection{RMS Prop}
\label{ss:rmsprop}

RMS Prop (Root Mean Square Propagation) is the optimization algorithm that I used to train most neural networks in this thesis. It deals with some of the challenges that vanilla gradient descent methods face. RMS Prop belongs to a family of gradient descent optimization algorithms that use momentum and/or adaptive learning rates. A more detailed discussion of these methods can be found in \cite{ruder2016overview}. Herein, I am using RMS Prop without further discussion. 

In RMS Prop, the learning rate is adapted for every single parameter in the parameter vector $\theta$. The idea is to divide the learning rate for a weight by a running average of the magnitudes of recent gradients for that weight \citep{rmsprop}. This running average is computed by:

$$
v_{t+1}(\theta) = \gamma \ v_t(\theta) + (1 - \gamma) \nabla \mathcal{L}_i(\theta)^2
$$

where $v_t$ is the moving average at step $t$ and $\gamma$ is the momentum rate, or forgetting factor. The parameter vector is then updated as follows:

$$
\theta_{t+1} = \theta_t - \frac{\eta}{\sqrt{v_{t+1}(\theta_t)}} \nabla \mathcal{L}_i (\theta_t)
$$

In my implementations, I am using the recommended values $\gamma = 0.9$ and $\eta = 0.001$ \citep{ruder2016overview} and \citep{rmsprop}.

\subsection{Futher Results}
\label{ss:appendix-results}

\begin{table}[h]
\begin{center}
\begin{tabular}{| c | c | c | c | c |}
  \hline
  Layers & Epochs & Time & Train acc & Test acc\\
  \hline
  5 x 750 & 50 / 27 & \textbf{266.9s} / 493.1s & \textbf{100.00\%} / 97.58\% & \textbf{97.67\%} / 96.38\%\\
  2 x 512 & 33 / 28 & \textbf{92.3s} / 123.3s & \textbf{99.95\%} / 98.53\% & 97.42\% / \textbf{97.57}\%\\
  3 x 850 & 24 / 21 & \textbf{133.1s} / 284.16s & \textbf{99.78\%} / 98.36\% & \textbf{97.36\%} / 96.97\%\\
  \hline 
\end{tabular}
\end{center}
\caption{Network performances when trained using forward thinking (left values) and backpropagation (right values).}
\label{tbl:ft-backprop-appendix}
\end{table}

  \bibliographystyle{apalike}
  \bibliography{bibliography/arxiv,bibliography/elsevier_nn,bibliography/ieee,bibliography/jmlr,bibliography/nips,bibliography/other_article,bibliography/other_book,bibliography/other_collection,bibliography/other_misc,bibliography/other_proceeding,bibliography/other_thesis}

\begin{thebibliography}{}

\bibitem[Bartlett et~al., 1999]{bartlett1999almost}
Bartlett, P.~L., Maiorov, V., and Meir, R. (1999).
\newblock {Almost Linear VC Dimension Bounds for Piecewise Polynomial
  Networks}.
\newblock In Kearns, M.~J., Solla, S.~A., and Cohn, D.~A., editors, {\em
  {Advances in Neural Information Processing Systems 11}}, pages 190--196. MIT
  Press.

\bibitem[Bergstra and Bengio, 2012]{bergstra2012random}
Bergstra, J. and Bengio, Y. (2012).
\newblock {Random Search for Hyper-Parameter Optimization}.
\newblock {\em {The Journal of Machine Learning Research}}, 13(Feb):281--305.

\bibitem[Bianchini and Scarselli, 2014]{bianchini2014complexity}
Bianchini, M. and Scarselli, F. (2014).
\newblock {On the Complexity of Shallow and Deep Neural Network Cassifiers}.
\newblock In {\em European Symposium on Artificial Neural Networks}, volume~22,
  pages 371--376.

\bibitem[Choromanska et~al., 2015]{lossmln}
Choromanska, A., Henaff, M., Mathieu, M., Arous, G.~B., and LeCun, Y. (2015).
\newblock {The Loss Surfaces of Multilayer Networks}.
\newblock In {\em {Proceedings of the 18th International Conference on
  Artificial Intelligence and Statistics}}, volume~38 of {\em JMLR: W\&CP},
  pages 192--204. JMLR.org.

\bibitem[Conneau et~al., 2016]{textclass}
Conneau, A., Schwenk, H., Barrault, L., and Lecun, Y. (2016).
\newblock {Very Deep Convolutional Networks for Text Classification}.
\newblock {\em arXiv preprint arXiv:1606.01781}.

\bibitem[Cortes et~al., 2017]{cortes2017adanet}
Cortes, C., Gonzalvo, X., Kuznetsov, V., Mohri, M., and Yang, S. (2017).
\newblock {Adanet: Adaptive Structural Learning of Artificial Neural Networks}.
\newblock In {\em {Proceedings of the 34th International Conference on Machine
  Learning}}, volume~70, pages 874--883. JMLR.

\bibitem[Eldan and Shamir, 2016]{eldan2016power}
Eldan, R. and Shamir, O. (2016).
\newblock {The Power of Depth for Feedforward Neural Networks}.
\newblock In {\em Conference on Learning Theory}, volume~49, pages 907--940.

\bibitem[Elsken et~al., 2019]{elsken2018neural}
Elsken, T., Metzen, J.~H., and Hutter, F. (2019).
\newblock {Neural Architecture Search: A Survey}.
\newblock {\em {The Journal of Machine Learning Research}}, 20(55):1--21.

\bibitem[Fahlman, 1991]{recascor}
Fahlman, S.~E. (1991).
\newblock {The Recurrent Cascade-Correlation Architecture}.
\newblock In Lippmann, R.~P., Moody, J.~E., and Touretzky, D.~S., editors, {\em
  {Advances in Neural Information Processing Systems 3}}, pages 190--196.
  Morgan-Kaufmann.

\bibitem[Fahlman and Lebiere, 1990]{cascor}
Fahlman, S.~E. and Lebiere, C. (1990).
\newblock {The Cascade-Correlation Learning Architecture}.
\newblock In Touretzky, D.~S., editor, {\em {Advances in Neural Information
  Processing Systems 2}}, pages 524--532. Morgan-Kaufmann.

\bibitem[Frean, 1990]{frean1990upstart}
Frean, M. (1990).
\newblock {The Upstart Algorithm: A Method for Constructing and Training
  Feedforward Neural Networks}.
\newblock {\em Neural Computation}, 2(2):198--209.

\bibitem[Goodfellow et~al., 2016]{deeplearning}
Goodfellow, I., Bengio, Y., and Courville, A. (2016).
\newblock {\em {Deep Learning}}.
\newblock MIT press.

\bibitem[Hanson, 1990]{hanson1990meiosis}
Hanson, S.~J. (1990).
\newblock {Meiosis Networks}.
\newblock In Touretzky, D.~S., editor, {\em {Advances in Neural Information
  Processing Systems 2}}, pages 533--541. Morgan-Kaufmann.

\bibitem[Harvey, 2017]{eannimplementation}
Harvey, M. (2017).
\newblock {Let’s evolve a neural network with a genetic algorithm — code
  included}.

\bibitem[Hassibi et~al., 1994]{hassibi1994optimal}
Hassibi, B., Stork, D.~G., and Wolff, G. (1994).
\newblock {Optimal Brain Surgeon: Extensions and Performance Comparisons}.
\newblock In Cowan, J.~D., Tesauro, G., and Alspector, J., editors, {\em
  {Advances in Neural Information Processing Systems 6}}, pages 263--270.
  Morgan-Kaufmann.

\bibitem[Hassibi et~al., 1993]{hassibi1993optimal}
Hassibi, B., Stork, D.~G., and Wolff, G.~J. (1993).
\newblock Optimal brain surgeon and general network pruning.
\newblock In {\em IEEE International Conference on Neural Networks}, volume~1,
  pages 293--299. IEEE.

\bibitem[He et~al., 2016]{he2016deep}
He, K., Zhang, X., Ren, S., and Sun, J. (2016).
\newblock {Deep Residual Learning for Image Recognition}.
\newblock In {\em {Proceedings of the IEEE Conference on Computer Vision and
  Pattern Recognition}}, volume~1, pages 770--778.

\bibitem[Hettinger et~al., 2017]{forwardthinking}
Hettinger, C., Christensen, T., Ehlert, B., Humpherys, J., Jarvis, T., and
  Wade, S. (2017).
\newblock {Forward Thinking: Building and Training Neural Networks One Layer at
  a Time}.
\newblock {\em arXiv preprint arXiv:1706.02480}.

\bibitem[Hinton, 2012]{hinton2012practical}
Hinton, G.~E. (2012).
\newblock {A Practical Guide to Training Restricted Boltzmann Machines}.
\newblock In {\em {Neural Networks: Tricks of the Trade}}, pages 599--619.
  Springer.

\bibitem[Hornik et~al., 1989]{mlpuniversalapprox}
Hornik, K., Stinchcombe, M., and White, H. (1989).
\newblock {Multilayer Feedforward Networks Are Universal Approximators}.
\newblock {\em Neural Networks}, 2(5):359--366.

\bibitem[Keras, 2019]{kerasmnist}
Keras (2019).
\newblock {Simple Deep Neural Network on the MNIST Dataset}.

\bibitem[Krizhevsky et~al., 2017]{imageclass}
Krizhevsky, A., Sutskever, I., and Hinton, G.~E. (2017).
\newblock {ImageNet Classification with Deep Convolutional Neural Networks}.
\newblock {\em Commununity of the Association of Computing Machinery},
  60(6):84--90.

\bibitem[Krogh and Hertz, 1992]{krogh1992simple}
Krogh, A. and Hertz, J.~A. (1992).
\newblock {A Simple Weight Decay Can Improve Generalization}.
\newblock In Moody, J.~E., Hanson, S.~J., and Lippmann, R.~P., editors, {\em
  {Advances in Neural Information Processing Systems 4}}, pages 950--957.
  Morgan-Kaufmann.

\bibitem[Kumar et~al., 2016]{question}
Kumar, A., Irsoy, O., Ondruska, P., Iyyer, M., Bradbury, J., Gulrajani, I.,
  Zhong, V., Paulus, R., and Socher, R. (2016).
\newblock {Ask Me Anything: Dynamic Memory Networks for Natural Language
  Processing}.
\newblock In {\em {Proceedings of the 33rd International Conference on Machine
  Learning}}, volume~48, pages 1378--1387. JMLR.

\bibitem[Larochelle et~al., 2007]{larochelle2007empirical}
Larochelle, H., Erhan, D., Courville, A., Bergstra, J., and Bengio, Y. (2007).
\newblock {An Empirical Evaluation of Deep Architectures on Problems with Many
  Factors of Variation}.
\newblock In {\em {Proceedings of the 24th International Conference on Machine
  Learning}}, pages 473--480. ACM.

\bibitem[LeCun and Bengio, 1998]{cnn}
LeCun, Y. and Bengio, Y. (1998).
\newblock {Convolutional Networks for Images, Speech, and Time Series}.
\newblock In Arbib, M.~A., editor, {\em {The Handbook of Brain Theory and
  Neural Networks}}, pages 255--258. MIT Press.

\bibitem[LeCun et~al., 1998]{mnist}
LeCun, Y., Bottou, L., Bengio, Y., Haffner, P., et~al. (1998).
\newblock {Gradient-Based Learning Applied to Document Recognition}.
\newblock {\em Proceedings of the IEEE}, 86(11):2278--2324.

\bibitem[LeCun et~al., 1990]{lecun1990optimal}
LeCun, Y., Denker, J.~S., and Solla, S.~A. (1990).
\newblock {Optimal Brain Damage}.
\newblock In Touretzky, D.~S., editor, {\em {Advances in Neural Information
  Processing Systems 2}}, pages 598--605. Morgan-Kaufmann.

\bibitem[LeCun et~al., 2012]{lecun2012efficient}
LeCun, Y.~A., Bottou, L., Orr, G.~B., and M{\"u}ller, K.-R. (2012).
\newblock {Efficient Backprop}.
\newblock In {\em {Neural Networks: Tricks of the Trade}}, pages 9--48.
  Springer.

\bibitem[Levin et~al., 1994]{levin1994fast}
Levin, A.~U., Leen, T.~K., and Moody, J.~E. (1994).
\newblock {Fast Pruning Using Principal Components}.
\newblock In Cowan, J.~D., Tesauro, G., and Alspector, J., editors, {\em
  {Advances in Neural Information Processing Systems 6}}, pages 35--42.
  Morgan-Kaufmann.

\bibitem[Levine et~al., 2016]{robotics}
Levine, S., Finn, C., Darrell, T., and Abbeel, P. (2016).
\newblock {End-To-End Training of Deep Visuomotor Policies}.
\newblock {\em {The Journal of Machine Learning Research}}, 17(1):1334--1373.

\bibitem[Littmann and Ritter, 1992]{littmann1992cascade}
Littmann, E. and Ritter, H. (1992).
\newblock Cascade network architectures.
\newblock In {\em [Proceedings 1992] IJCNN International Joint Conference on
  Neural Networks}, volume~2, pages 398--404. IEEE.

\bibitem[Littmann and Ritter, 1993]{littmann1993generalization}
Littmann, E. and Ritter, H. (1993).
\newblock {Generalization Abilities of Cascade Network Architecture}.
\newblock In Hanson, S.~J., Cowan, J.~D., and Giles, C.~L., editors, {\em
  {Advances in Neural Information Processing Systems 5}}, pages 188--195.
  Morgan-Kaufmann.

\bibitem[Maass et~al., 1994]{maass1994comparison}
Maass, W., Schnitger, G., and Sontag, E.~D. (1994).
\newblock {A Comparison of the Computational Power of Sigmoid and Boolean
  Threshold Circuits}.
\newblock In {\em {Theoretical Advances in Neural Computation and Learning}},
  pages 127--151. Springer.

\bibitem[Mezard and Nadal, 1989]{mezard1989learning}
Mezard, M. and Nadal, J.-P. (1989).
\newblock {Learning in Feedforward Layered Networks: The Tiling Algorithm}.
\newblock {\em Journal of Physics A: Mathematical and General}, 22(12):2191.

\bibitem[Mozer and Smolensky, 1989]{skeletonization}
Mozer, M.~C. and Smolensky, P. (1989).
\newblock {Skeletonization: A Technique for Trimming the Fat from a Network via
  Relevance Assessment}.
\newblock In Touretzky, D.~S., editor, {\em {Advances in Neural Information
  Processing Systems 1}}, pages 107--115. Morgan-Kaufmann.

\bibitem[Nowlan and Hinton, 1992]{nowlan1992simplifying}
Nowlan, S.~J. and Hinton, G.~E. (1992).
\newblock {Simplifying Neural Networks by Soft Weight-Sharing}.
\newblock {\em Neural Computation}, 4(4):473--493.

\bibitem[Poole et~al., 2016]{poole2016exponential}
Poole, B., Lahiri, S., Raghu, M., Sohl-Dickstein, J., and Ganguli, S. (2016).
\newblock {Exponential Expressivity in Deep Neural Networks Through Transient
  Chaos}.
\newblock In Lee, D.~D., Sugiyama, M., Luxburg, U.~V., Guyon, I., and Garnett,
  R., editors, {\em {Advances in Neural Information Processing Systems 29}},
  pages 3360--3368. Curran Associates, Inc.

\bibitem[Prechelt, 1997]{prechelt1997investigation}
Prechelt, L. (1997).
\newblock {Investigation of the Cascor Family of Learning Algorithms}.
\newblock {\em Neural Networks}, 10(5):885--896.

\bibitem[Raghu et~al., 2017]{raghu2017expressive}
Raghu, M., Poole, B., Kleinberg, J., Ganguli, S., and Dickstein, J.~S. (2017).
\newblock {On the Expressive Power of Deep Neural Networks}.
\newblock In {\em {Proceedings of the 34th International Conference on Machine
  Learning}}, volume~70 of {\em JMLR: W\&CP}, pages 2847--2854. JMLR.org.

\bibitem[Real et~al., 2018]{real2018regularized}
Real, E., Aggarwal, A., Huang, Y., and Le, Q.~V. (2018).
\newblock Regularized evolution for image classifier architecture search.
\newblock {\em arXiv preprint arXiv:1802.01548}.

\bibitem[Reed, 1993]{reed1993pruning}
Reed, R. (1993).
\newblock Pruning algorithms - a survey.
\newblock {\em IEEE Transactions on Neural Networks}, 4(5):740--747.

\bibitem[Ruder, 2016]{ruder2016overview}
Ruder, S. (2016).
\newblock {An Overview of Gradient Descent Optimization Algorithms}.
\newblock {\em arXiv preprint arXiv:1609.04747}.

\bibitem[Sermanet et~al., 2013]{overfeat}
Sermanet, P., Eigen, D., Zhang, X., Mathieu, M., Fergus, R., and LeCun, Y.
  (2013).
\newblock {Overfeat: Integrated Recognition, Localization and Detection Using
  Convolutional Networks}.
\newblock {\em arXiv preprint arXiv:1312.6229}.

\bibitem[Sietsma, 1988]{sietsma1988neural}
Sietsma, J. (1988).
\newblock {Neural Net Pruning - Why and How}.
\newblock In {\em {Proceedings of International Conference on Neural
  Networks}}, volume~1, pages 325--333.

\bibitem[Silver et~al., 2018]{alphazero}
Silver, D., Hubert, T., Schrittwieser, J., Antonoglou, I., Lai, M., Guez, A.,
  Lanctot, M., Sifre, L., Kumaran, D., Graepel, T., et~al. (2018).
\newblock {A General Reinforcement Learning Algorithm that Masters Chess,
  Shogi, and Go through Self-Play}.
\newblock {\em Science}, 362(6419):1140--1144.

\bibitem[Sjogaard, 1991]{sjogaard1991conceptual}
Sjogaard, S. (1991).
\newblock {\em {A Conceptual Approach to Generalisation in Dynamic Neural
  Networks}}.
\newblock PhD thesis, Aarhus University.

\bibitem[Srivastava et~al., 2014]{srivastava2014dropout}
Srivastava, N., Hinton, G., Krizhevsky, A., Sutskever, I., and Salakhutdinov,
  R. (2014).
\newblock {Dropout: A Simple Way to Prevent Neural Networks from Overfitting}.
\newblock {\em {The Journal of Machine Learning Research}}, 15(1):1929--1958.

\bibitem[Telgarsky, 2015]{telgarsky2015representation}
Telgarsky, M. (2015).
\newblock {Representation Benefits of Deep Feedforward Networks}.
\newblock {\em arXiv preprint arXiv:1509.08101}.

\bibitem[Tieleman and Hinton, 2012]{rmsprop}
Tieleman, T. and Hinton, G. (2012).
\newblock {Lecture 6.5-rmsprop}.

\bibitem[Vapnik and Chervonenkis, 2015]{vapnik2015uniform}
Vapnik, V.~N. and Chervonenkis, A.~Y. (2015).
\newblock {On the Uniform Convergence of Relative Frequencies of Events to
  Their Probabilities}.
\newblock In {\em {Measures of Complexity}}, pages 11--30. Springer.

\bibitem[Vinyals et~al., 2015]{captions}
Vinyals, O., Toshev, A., Bengio, S., and Erhan, D. (2015).
\newblock Show and tell: A neural image caption generator.
\newblock In {\em IEEE Conference on Computer Vision and Pattern Recognition},
  volume~28, pages 3156--3164.

\bibitem[Wang et~al., 1994]{wang1994procedure}
Wang, Z., Di~Massimo, C., Tham, M.~T., and Morris, A.~J. (1994).
\newblock {A Procedure for Determining the Topology of Multilayer Feedforward
  Neural Networks}.
\newblock {\em Neural Networks}, 7(2):291--300.

\bibitem[Waugh, 1994]{waugh1994dynamic}
Waugh, S. (1994).
\newblock {Dynamic learning algorithms}.
\newblock Department of Computer Science, University of Tasmania.

\bibitem[Wu et~al., 2016]{transl}
Wu, Y., Schuster, M., Chen, Z., Le, Q.~V., Norouzi, M., Macherey, W., Krikun,
  M., Cao, Y., Gao, Q., Macherey, K., et~al. (2016).
\newblock {Google's Neural Machine Translation System: Bridging the Gap Between
  Human and Machine Translation}.
\newblock {\em arXiv preprint arXiv:1609.08144}.

\bibitem[Wynne-Jones, 1992]{wynne1992node}
Wynne-Jones, M. (1992).
\newblock {Node Splitting: A Constructive Algorithm for Feed-Forward Neural
  Networks}.
\newblock In Moody, J.~E., Hanson, S.~J., and Lippmann, R.~P., editors, {\em
  {Advances in Neural Information Processing Systems 4}}, pages 1072--1079.
  Morgan-Kaufmann.

\bibitem[Yang and Honavar, 1998]{yang1991experiments}
Yang, J. and Honavar, V. (1998).
\newblock {Experiments with the Cascade-Correlation Algorithm}.
\newblock {\em Microcomputer Applications}, 17(2):40--46.

\bibitem[Yao, 1999]{evolvinganns}
Yao, X. (1999).
\newblock {Evolving Artificial Neural Networks}.
\newblock {\em {Proceedings of the IEEE}}, 87(9):1423--1447.

\bibitem[Zhang et~al., 2016]{zhang2016understanding}
Zhang, C., Bengio, S., Hardt, M., Recht, B., and Vinyals, O. (2016).
\newblock {Understanding Deep Learning Requires Rethinking Generalization}.
\newblock {\em arXiv preprint arXiv:1611.03530}.

\bibitem[Zoeller and Huber, 2019]{zoller2019survey}
Zoeller, M. and Huber, M. (2019).
\newblock Survey on automated machine learning.
\newblock {\em arXiv preprint arXiv:1904.12054}.

\bibitem[Zoph et~al., 2018]{Zoph_2018_CVPR}
Zoph, B., Vasudevan, V., Shlens, J., and Le, Q.~V. (2018).
\newblock {Learning Transferable Architectures for Scalable Image Recognition}.
\newblock In {\em {The IEEE Conference on Computer Vision and Pattern
  Recognition (CVPR)}}, volume~1.

\end{thebibliography}
  
\end{document}